\documentclass[journal]{IEEEtran}
\usepackage{longtable}
\usepackage[pdftex]{graphicx}
\usepackage{subfig}
\ifCLASSINFOpdf
\else
\fi
\begin{document}
%
\title{Estimating sex and age for forensic applications using machine learning based on facial measurements from frontal cephalometric landmarks}
%
%
%

\author{Lucas F. Porto, Laise N. Correia Lima, Ademir Franco, Donald M. Pianto, Carlos Eduardo Machado Palhares, Donald M. Pianto and Flavio de Barros Vidal\thanks{Lucas F. Porto, Flavio de B. Vidal are with the University of Brasilia, Brasilia-DF, Brazil.}\thanks{Laise N. Correia Lima is with the Federal University of Maranhao, Sao Luis-MA, Brazil.}\thanks{Ademir Franco is with KU Leuven, Kapucijnenvoer 7, block B, Leuven, Belgium.}\thanks{Carlos Eduardo M. Palhares is National Institute of Criminalistics, Technical Scientific Direction/Brazilian Federal Police, Brasilia-DF, Brazil.}\thanks{Manuscript received XXX; revised XXX, 2019.}}

\markboth{Journal of \LaTeX\ Class Files,~Vol.~14, No.~8, August~2015}%
{Shell \MakeLowercase{\textit{et al.}}: Bare Demo of IEEEtran.cls for IEEE Journals}
%



\maketitle

\begin{abstract}
Facial analysis permits many investigations some of the most important of which are craniofacial identification, facial recognition, and age and sex estimation. In forensics, photo-anthropometry describes the study of facial growth and allows the identification of patterns in facial skull development by using a group of cephalometric landmarks to estimate anthropological information. Previous works presented, as indirect applications, the use of photo-anthropometric measurements to estimate anthropological information such as age and sex. In several areas, automation of manual procedures has achieved advantages over and similar measurement confidence as a forensic expert. This manuscript presents an approach using photo-anthropometric indexes, generated from frontal faces cephalometric landmarks, to create an artificial neural network classifier that allows the estimation of anthropological information, in this specific case age and sex. The work is focused on four tasks: i) sex estimation over ages from 5 to 22 years old, evaluating the interference of age on sex estimation; ii) age estimation from photo-anthropometric indexes for four age intervals (1 year, 2 years, 4 years and 5 years); iii) age group estimation for thresholds of over 14 and over 18 years old; and; iv) the provision of a new data set, available for academic purposes only, with a large and complete set of facial photo-anthropometric points marked and checked by forensic experts, measured from over 18,000 faces of individuals from Brazil over the last 4 years. The proposed classifier obtained significant results, using this new data set, for the sex estimation of individuals over 14 years old, achieving accuracy values greater than 0.85 by the \textit{F}$_{1}$ measure. For age estimation, the accuracy results are 0.72 for measure with an age interval of 5 years. For the age group estimation, the measures of accuracy are greater than 0.93 and 0.83 for thresholds of 14 and 18 years, respectively.
\end{abstract}


%
\IEEEpeerreviewmaketitle

\section{Introduction}\label{sec-introduction}



Anthropological knowledge can be used to support forensic investigations of the deceased and the living~\cite{marquez2015overview}. In the first case, postmortem (PM) profiles of victims are reconstructed to narrow the number of comparisons between missing persons and unknown bodies, as described in~\cite{silva2013interrelationship}. The profiling process is carried out by retrieving information regarding sex, age, stature and ancestry from the deceased, especially from skeletal remains, and comparing them with antemortem (AM) data from the alleged victim~\cite{adserias2018forensic}. In order to increase reliability, the collected information is combined with AM and PM evidence obtained by primary means of human identification, namely fingerprint, dental and DNA analyses~\cite{interpol2018DVI-Lion}. 

On the other hand, forensic anthropology applied to the living usually relies on morphological and biometric information of victims and suspects of crimes registered in footage of closed-circuit television and photographs~\cite{cattaneo2012can,ratnayake2014juvenile}. The identification of children that suffered sexual exploitation, as well as their perpetrators, figures among the procedures requested by Law in the routines of medico-legal institutes~\cite{BORGES20189}. Over the last decade, requests for anthropological examination of the living became more common following an increasing trend in cybernetic crimes~\cite{cattaneo2009difficult}. This new outlook justified the need for developing advanced tools to support forensic casework~\cite{palhares2017}.

The photo-anthropometric analysis of the human face emerges in this context as an alternative tool for searching, collecting, and quantifying morphological features and using them for forensic purposes~\cite{flores2018comparative}. Working at the interface of forensic anthropology and computer science, this non-invasive and low-cost approach is founded on the registration of landmarks on photographs and the calculation of ratios between facial distances~\cite{BORGES20189,palhares2017,flores2018comparative}. The morphometric information retrieved from the human face can be used within a comparative basis, between reference and target persons, or in a reconstructive basis, where sexual dimorphism and age estimation of the living are performed~\cite{gonzales2018photoanthropometry}.

This study was designed with four aims for the use of photo-anthropometric data of the human face: to propose an automatic solution based on an artificial neural network to estimate anthropological information using photo-anthropometric indices (I); to test the diagnostic accuracy of the solution with cut-off points between male and females and threshold limits for the ages of 14 - related to sexual consent~\cite{zhu2017trends,carpenter2014harm}, and 18 - related  to legal majority~\cite{cericato2016correlating,machado2018effectiveness} (II); to analyze the correlation between sex and age using photo-anthropometric indices of the human face (III) and; provision, for academic purposes only, of a complete data set of facial photo-anthropometric points marked and checked by forensic experts, measured on over 18.000 faces of individuals from Brazil over the last 4 years (IV).

The manuscript is organized as follows: Section~\ref{sec:materialmethods} presents the proposed method and all the processing details for classification using a model based on artificial intelligence techniques. It also includes a description of the database and the inputs used in this manuscript. Section~\ref{sec:results} presents the experimental results. Finally, Section~\ref{sec:conclusion} presents the discussion, main conclusions and future works.

\section{Materials and Methods}\label{sec:materialmethods}

The main goal of this manuscript is to evaluate the use of photo-anthropometric data from human faces to create an automatic classifier based on an artificial neural network to estimate age and sex. In this Section we will describe the approach used to create an automatic solution using an artificial intelligence model and the details of the inputs and the proposed tests.

\subsection{Photos Database and Cephalometric Landmarks sets}\label{subsec:DataLandMark}

The principal proposed and used database is composed of photo-anthropometric index data from 18,000 frontal face photos of 18,000 different people from Brazil. All photos were acquired in accordance with ICAO 9303 normative~\cite{ISO19794-5} used for Machine Readable Travel Documents. All photos have been captured over a white background, stored at a resolution of 480x640 pixels, 24bits of color, with no glasses and with a natural expression. The photos are divided into male (9,000) and female (9,000) and 18 age groups (from 5 up to 22 years old), totaling 500 photos for each individual group (sex and age).

\begin{figure}
\centering
\includegraphics[width=250px]{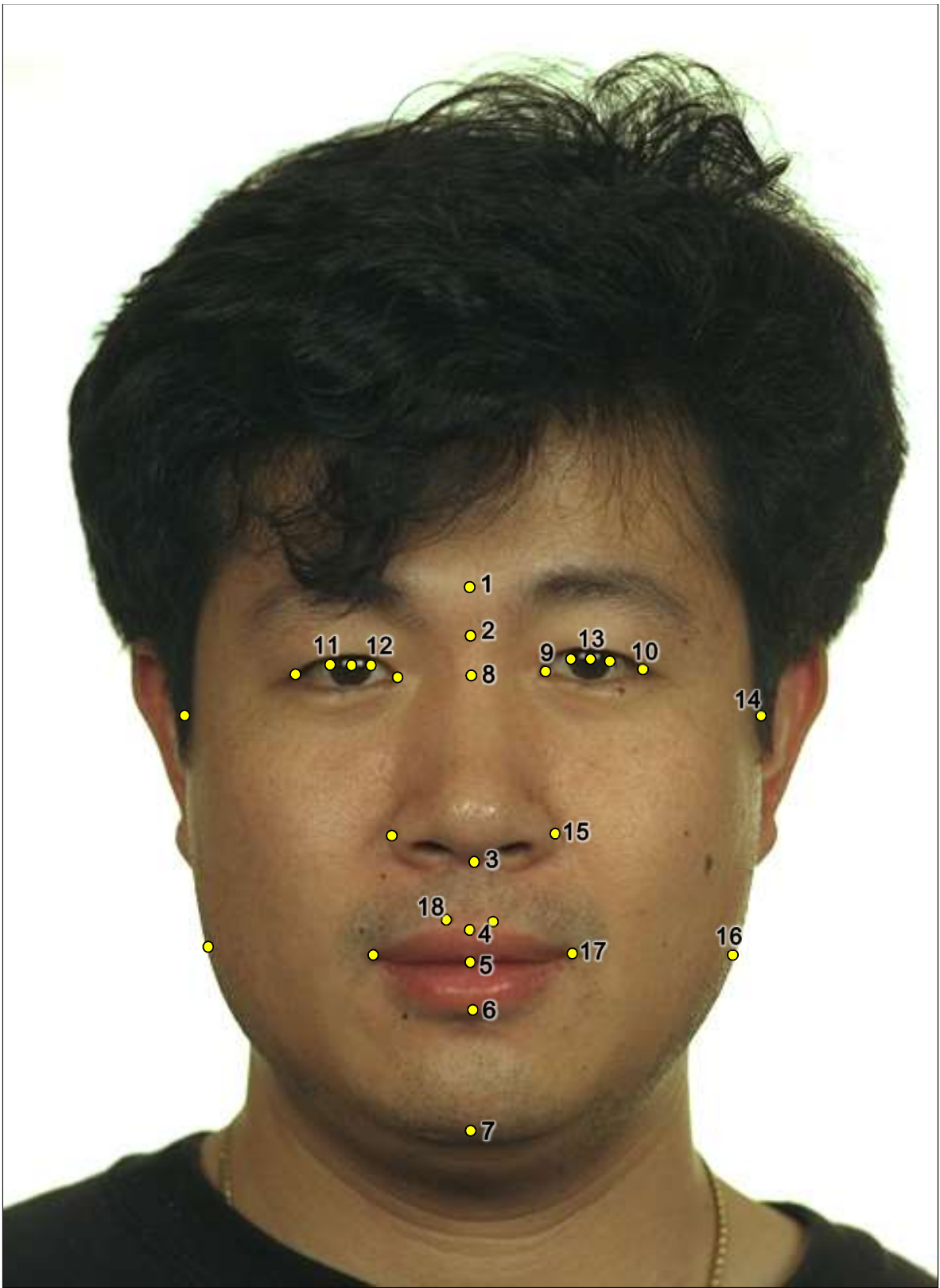}
\caption{All 28 cephalometric landmarks adopted in this work: 1.~\textit{Glabella} (g'); 2.~\textit{Nasion} (n'); 3.~\textit{Subnasale} (sn'); 4.~\textit{Labiale superius} (ls'); 5.~\textit{Stomion} (sto'); 6.~\textit{Labiale inferius} (li'); 7.~\textit{Gnathion} (gn'); 8.~\textit{Midnasal} (m'); 9.~\textit{Endocanthion} (en'); 10.~\textit{Exocanthion} (ex'); 11.~\textit{Iridion laterale} (il); 12.~\textit{Iridion mediale} (im); 13.~\textit{Pupil} (pu'); 14.~\textit{Zygion} (zy'); 15.~\textit{Alare} (al'); 16.~\textit{Gonion} (go'); 17.~\textit{Cheilion} (ch'); 18.~\textit{Crista philtri} (cph'), image adapted from~\protect\cite{phillips2000feret}.}
\label{fig:all_points}
\end{figure}

The cephalometric landmarks adopted in this manuscript, all of which were used to create the photo-anthropometric indexes, were described in~\cite{palhares2017} and~\cite{BORGES20189}. In this methodology, one expert manually located 28 cephalometric landmarks in each of a 1000 pre-training images using a SAFF-2D software, applying an identification methodology proposed in~\cite{flores2018comparative,flores2017manual,flores2014master}. All 18,000 faces were processed and all facial landmarks were marked and checked following the methodology developed in~\cite{PrePrintFSI2019}. The following is a complete list of cephalometric landmarks which were identified, according to Caple and Stephan standard nomenclature~\cite{stephanStandards16}:~\textit{Entocanthion} (en'),~\textit{Exocanthion} (ex'),~\textit{Iridion laterale} (il),~\textit{Iridion mediale} (im),~\textit{Pupil} (pu'),~\textit{Zygion} (zy'),~\textit{Alare} (al'),~\textit{Gonion} (go') and~\textit{Cheilion} (ch'),~\textit{Crista philtri} (cph') bilateral landmarks. The remaining can be found on face's midline as:~\textit{Glabella} (g'),~\textit{Nasion} (n'),~\textit{Subnasale} (sn'),~\textit{Labiale superius} (ls'),~\textit{Stomion} (sto'),~\textit{Labiale inferius} (li'),~\textit{Gnathion} (gn'),~\textit{Midnasal} (m')~\cite{brown2004survey}. Figure~\ref{fig:all_points} shows all the landmarks used and their locations on the face.

\subsection{Photo-anthropometric indexes}\label{subsec:pai}



Using the cephalometric landmarks detailed in Subsection~\ref{subsec:DataLandMark}, the authors of~\cite{palhares2017} and~\cite{BORGES20189} present details about 10 and 40 facial~\textbf{photo-anthropometric indexes}~(PAIs) respectively. They are facial measures which permit the explanation of people's anthropometric information, specifically age and sex, by only using facial growth information.

The main feature of the PAIs is the use of the mean iris diameter as a proportional factor defined as the \textit{iris ratio} to solve scale and calibration issues (e.g. different metrics relations to the pixel size and to the intrinsic parameters of the capture device which was used). The \textit{iris ratio} is essentially the Euclidean distance defined by a function $d(p,q)$ from \textit{Iridion laterale} (il) and \textit{Iridion mediale} (im) of both eyes. The \textit{iris ratio}  is detailed in Equation~\ref{eq:iris}. When a facial measurement generated by the Euclidean distance between two cephalometric landmarks is divided by the iris ratio the result is a stable growing factor which solves the image scale problem and removes effects of varying landmark positions over the face.

\begin{equation}\label{eq:iris}
\textit{iris\_ratio} = \frac{d(im_L,il_L)+d(im_R,il_R)}{2}
\end{equation}



Using a specialized software to automatically extract the cephalometric landmarks, we applied the methodology (\textit{iris ratio}) proposed in~\cite{palhares2017}, combining all of the 28 cephalometric landmarks creating 208 PAIs. The result is stored in a comma-separated values file (CSV) composed of the photo-anthropometry data, including 208 PAIs per image, labeled by sex and age for the 18.000 images used to build the data set. More details about the generated data set will be described in Section \ref{subsec:statistical-tests}. The complete description of all 208 PAIs can be found in the supplementary material in Section~\ref{appendix:208_PAI}. 

\subsection{Experimental set-up and evaluation metrics}\label{subsec:experimental}

All the proposed tests were executed on an Intel AI DevCloud~\cite{intelDev}, a cloud framework for applications in artificial intelligence powered by Intel, running the Keras API version 2.2.0~\cite{chollet2015keras} and Tensorflow version 1.8~\cite{tensorflow2015-whitepaper}. During the training process we adopted an artificial neural network (ANN) model based on a multilayer perceptron (MLP)~\cite{gardner1998artificial}, using one dense (fully connected) layer with 128 neurons with the Adamax optimizer from Adam~\cite{kingma2014adam}. Table~\ref{tab:tableMLP} presents the detailed MLP structure used to execute the tests.

\begin{table}[htbp]
\centering
  \caption{The detailed multilayer perceptron structure used to execute the tests.}
\begin{tabular}{ll}
\hline
\textbf{\# Input layer}  & 209 (208 PAIs + sex attribute) \\
\textbf{\# Hidden layer} & 1 \\
\textbf{\# Neuron in hidden layer} & 128 \\
\textbf{\# Epochs} & 500 \\
\textbf{Activation function in hidden layer} & Sigmoid \\
\textbf{Activation function in output layer} & Softmax \\
\textbf{Learning rate} & 0.01 \\
\textbf{Momentum} & 0.9 \\ 
\textbf{Optimizer} & Adamax \\  
\hline
\end{tabular}
  \label{tab:tableMLP}%
\end{table}

In a classification process, all results require a test accuracy analysis and a traditional measure is the \textit{F}$_{1}$ score (also known as F-score or F-measure)~\cite{fscoremeasure}. This accuracy metric is composed of four parameters: True Positives ($n_{tp}$), True Negatives ($n_{tn}$), False Positives ($n_{fp}$) and False Negatives ($n_{fn}$). At the end of the estimation process, we evaluated the estimate as a true positive ($n_{tp}$) when the classifier agreed with the anthropological data (age/sex) for the validation data and as $n_{fp}$ (false positive) when the estimate (age/sex) disagreed. We have used the \textit{F}$_{1}$ score method shown in Equation \ref{eq:fscore}, 

\begin{equation}\label{eq:fscore}
\textit{F}_{1} = 2 * \left( \frac{precision * recall}{precision - recall} \right)
\end{equation}

where the variables $precision$ and $recall$ are defined as $precision = n_{tp}/(n_{tp} + n_{fp})$ and $recall = n_{tp}/(n_{tp} + n_{fn})$, respectively. The  \textit{F}$_{1}$ score metric varies between 1 (best) and 0 (worst). To evaluate each test in detail, we adopted the confusion matrix method~\cite{Provost98onapplied}. The confusion matrix allows the comparison of each ``predicted label" with the ``true label" resulting in a table which presents the classifier's behavior, providing a more complete visualization than the \textit{F}$_{1}$ score.

As described above, we propose the use of 208 PAIs per photo and provide them as input to an automatic classifier to estimate anthropological information. In order to evaluate the expected performance of the classifier model over the proposed data set, a \textit{k-fold} cross-validation was used in the testing process~\cite{hastie_09,kohavi1995study}. In our case, a ten-fold cross-validation procedure was used, randomly separating our data set five times with $90\%$ of photos for training the classifiers and $10\%$ of photos for testing.

To evaluate the PAI methodology for anthropological estimation, we defined 3 groups of tests. The first one, \textbf{Group A} relates to the estimation of sex, the second one, \textbf{Group B}, to the estimation of age and \textbf{Group C} to the estimation of the age group. These tests are described as follows:

\begin{itemize}
    \item \textbf{Group A - Sex estimation}: The group is composed of two test sets where one classifier was trained with age as an input and the other classifier avoiding\footnote{When the term ``avoiding'' is used throughout the manuscript, it means information from a specific class was not used in the proposed tests.} it, working similarly to an adversarial neural network~\cite{NIPS2014_5423}.
    
    \item \textbf{Group B - Age estimation}: This group is composed of 4 tests to evaluate age estimation using only the PAI data. For all these tests, we focus on identifying if the sex data can provide improvements over (or interfere with) the age estimation using only the facial measurements.
    
    \item \textbf{Group C - Group age estimation}: In this group two classifiers are proposed in order to evaluate if the facial measurements are able to identify if an individual belongs to a specific age group. The classifiers analyze the thresholds of 14 and 18 years old. The first one identifies if the person is 14 years old or more and the second one if the person is 18 years old or older.
\end{itemize}

As stated above, in \textbf{Group A} the estimation of sex is analyzed. Table~\ref{tab:table01} presents the structure of the tests for Group A. In Test 1, we trained the classifier separately for each age group, totaling 17 individual tests from 5 up to 22 years old. This test was developed to try to evaluate the classification process applied on each age range using only the PAI data. In Test 2 we used all the data observations (avoiding the age information) to evaluate the sex classification over all age groups.

\begin{table}[htbp]
  \centering
  \caption{Group A: Structure of tests for sex estimation.}
    \begin{tabular}{cccc}
    \hline
     Test & Target & Age as input & Qty. Test \\ \hline \hline
    1 & Sex   & Yes   & 17 \\
    2 & Sex   & No    & 1 \\ \hline
    \end{tabular}%
  \label{tab:table01}%
\end{table}%


Meanwhile in \textbf{Group B} the age estimation process is analyzed. Table~\ref{tab:table02} presents the test structure of the Group B tests. We defined 4 tests to evaluate the classification process for age estimation using the PAI data. For each test we analyzed the impact of sex information on the estimation process.

\begin{table}[htbp]
  \centering
  \caption{Group B: Structure of tests for age estimation.}
    \begin{tabular}{ccc}
     \hline
         Test & Target & Sex as input \\ \hline  \hline
    1 & 6,7,8,9,10,11,12,13,14,15,16,17,18,19,20,21,22 & No/Female/Male \\
    2 & 6,8,10,12,14,16,18 ,20,22 & No/Female/Male \\
    3 & 6,10,14,18,22 & No/Female/Male \\
    4 & 5,10,15,20 & No/Female/Male \\
     \hline
    \end{tabular}%
  \label{tab:table02}%
\end{table}%

For each proposed test, we trained three different classifiers. Two by splitting the PAI data by sex (female and male) and the other classifier using all the data observations in the same classification process. The 4 tests in \textbf{Group B} are described as follow:

\begin{itemize}
    \item Test 1: We selected the PAI data from 6 to 22 years old at age intervals of 1 year. In this test we are focused on evaluating the classifier accuracy for age estimation with a total of 17 age classes as output.
    
    \item Test 2: We selected the PAI data from 6 to 22 years old at age intervals of 2 years. In this test, as in Test 1, we evaluate the classifier accuracy for age estimation with 9 age classes as output: 6, 8, 10, 12, 14, 16, 18, 20 and 22.
    
    \item Test 3: We selected the PAI data from 6 to 22 years old at age intervals of 4 years. As above, we are evaluating the classifier accuracy for age estimation with a total of 5 age class outputs: 6, 10, 18 and 22.
    
    \item Test 4: We selected the PAI data from 5 to 20 years old at age intervals of 5 years. As in all the previous tests of this group, we are interested in evaluating the classifier accuracy for age estimation with a total of 4 output classes: 5, 10, 15 and 20.
\end{itemize}

Finally, for \textbf{Group C}, Table~\ref{tab:table03} presents the test structure for age group classification, with thresholds of 14 and 18 years. In these tests we evaluated if the PAIs can be used to estimate if the person is older/under 14 years old or older/under 18 years old. We adopted the same test procedures as presented in \textbf{Group B}. In this case we evaluated if the sex data affects the age group estimation process. For both tests we trained classifiers for females, for males and for all observations without the sex information as input.

\begin{table}[htbp]
  \centering
  \caption{Group C: Age group, structure of tests for threshold of 14 and 18 years.}
    \begin{tabular}{ccc}
    \hline
       Test & Target & Sex as input  \\  \hline  \hline
    1 & Older/under 14 years & No/Female/Male \\
    2 & Older/under 18 years & No/Female/Male \\
    \hline
    \end{tabular}%
  \label{tab:table03}%
\end{table}%

\subsection{Descriptive Statistical Analyses of the Data Set}\label{subsec:statistical-tests}

\begin{figure}[!htpb]
\centering
\includegraphics[width=255px]{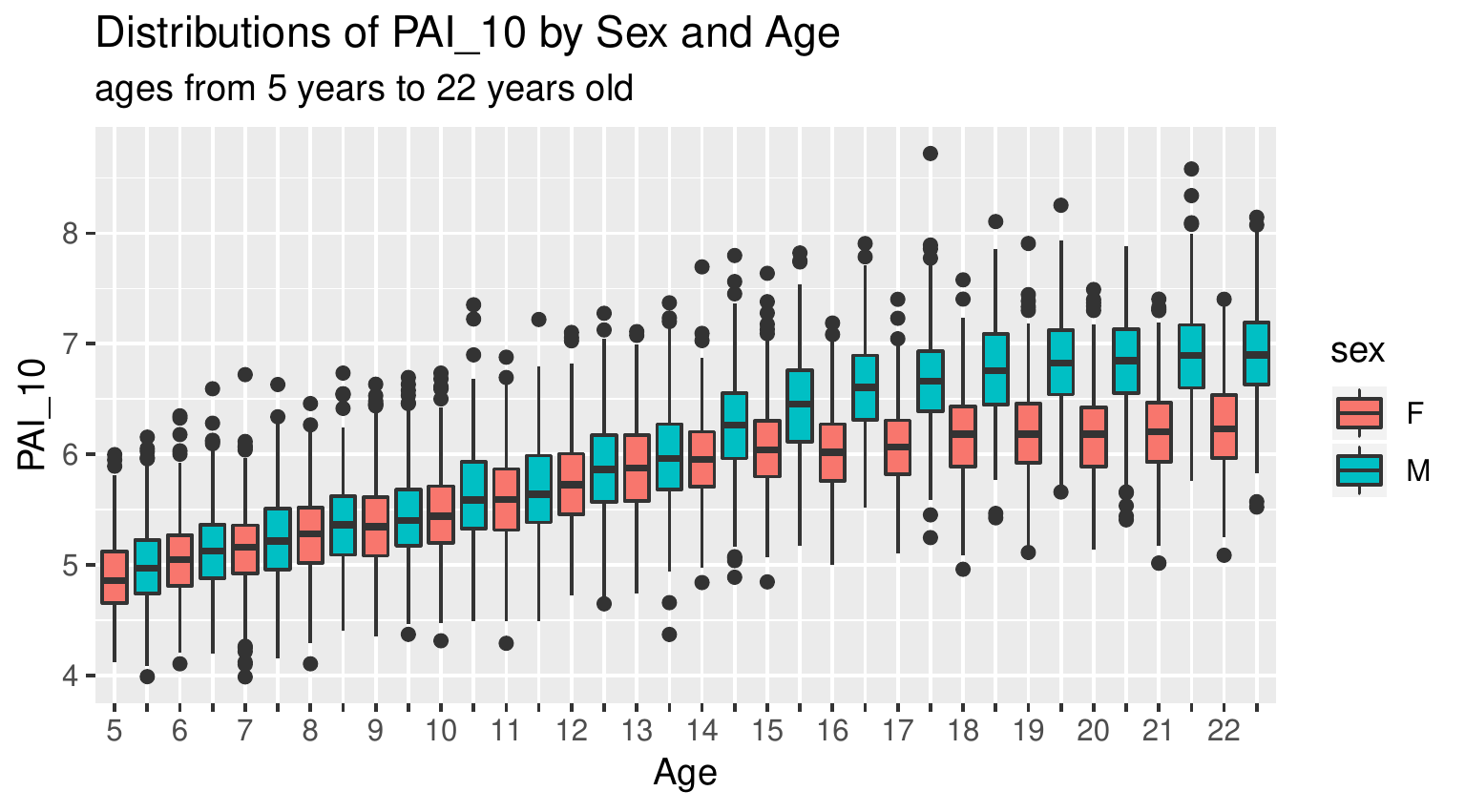}
\caption{Boxplots of PAI-10 (wing of the nose - chin) for each sex and age group}
\label{fig:PAI_10}
\end{figure}

In this section we undertake a descriptive statistical exploration of the new data set which serves as input to the proposed machine learning model\footnote{The data set and the developed machine learning model will be available to download soon, after the blind review process is completed, in attendance of the journal's submission criteria.}. As described above, the full data set contains 500 observations on 208 variables for each category of Age (18 levels) and Sex (2 levels), yielding $18,000$ observations in total. Given the large number of variables and categories, it is not a viable option to describe all the variable distributions in all the categories. Therefore, we performed a visual and detailed analysis of boxplots of the distributions of all the variables in all the 36 categories. An example for the variable PAI-10 (wing of the nose - chin) can be found in Figure~\ref{fig:PAI_10}. Males present larger values than females and the differences increase with age. Similar figures for all 208 variables can be found in the supplementary material\footnote{All boxplots are available in the specific section of the Supplementary Material files (File PAIs\_sex\_age\_boxplots\_supplemental.pdf).}. 

We performed Shapiro-Wilk Normality Tests~\cite{wilk1965test} in each age-sex group for each variable. The null hypothesis for this test is that the data are normally distributed. $P$ values lower than $0.01$ were considered indications of significant deviations from normality. Three variables were non-normal in all age-sex groups: PAI-154 (diameter of the iris), PAI-160 (lateral iris - pupil), and PAI-171 (medial iris - pupil, same side). For the other variables, approximately $12\%$ of the tests rejected normality (about two age-sex groups per variable). However, given the large number of observations in each category and the presence of many outliers\footnote{See Figure~\ref{fig:PAI_10}, for example, where outliers can be identified as the points well above or well below the boxes in the boxplot.}, both of which increase the probability of rejecting normality, this suggests that much of the data is normally distributed and that analysis techniques that depend on normality can be used, albeit with caution.

\begin{figure}[!htpb]
\centering
\includegraphics[width=255px]{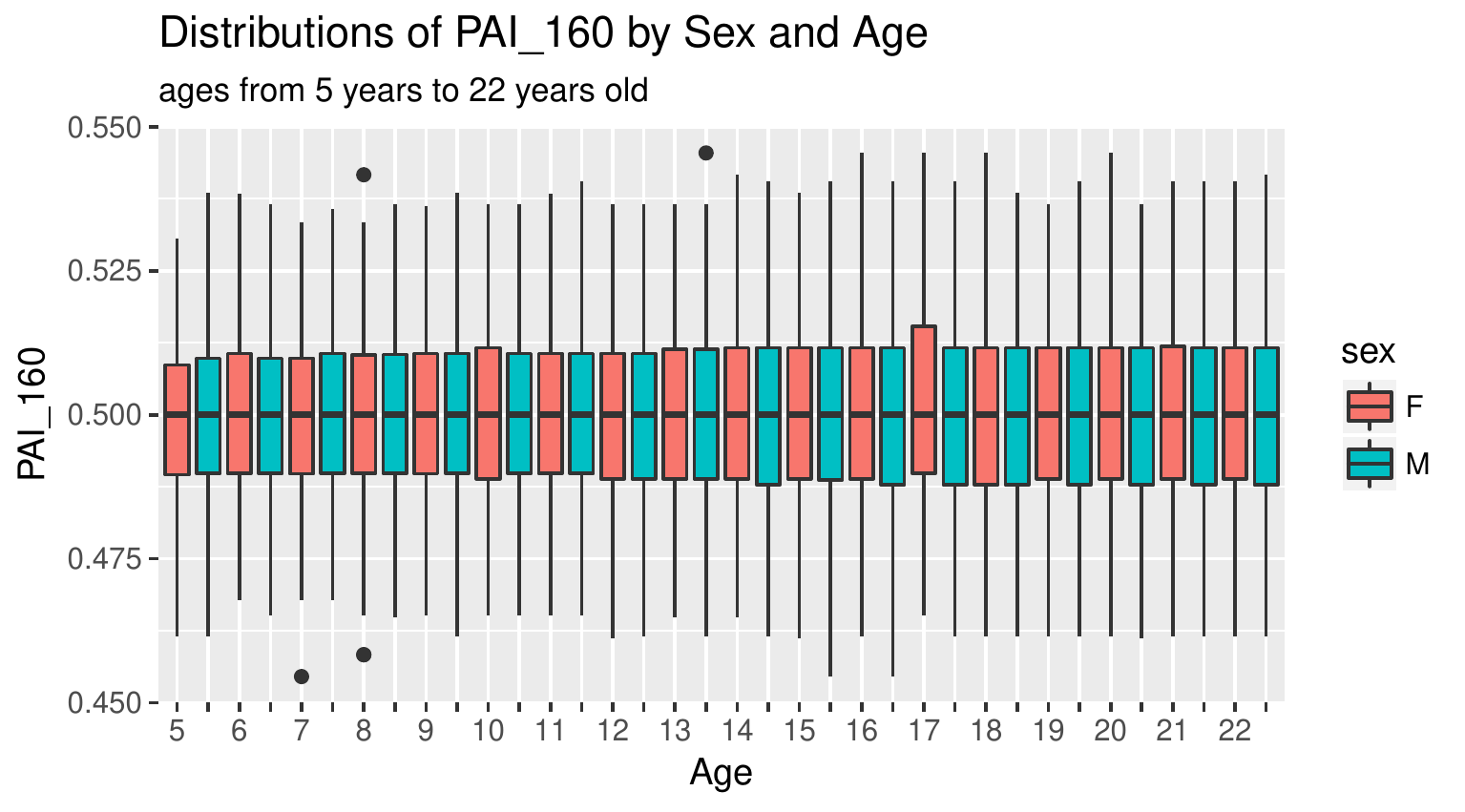}
\caption{Boxplots of PAI-160 (lateral iris - pupil) for each sex and age group}
\label{fig:PAI_160}
\end{figure}

Two-way analyses of variance (ANOVA)~\cite{kutner2005applied} were performed on each PAI to assess statistically significant changes in facial parameters in response to the factors sex, age, and the interaction between sex and age. $P$ values lower than $0.01$ were considered significant. For each of the 208 PAIs both sex and age were significant factors. The interaction between sex and age was not significant for only 4 of the 208 PAIs, specifically PAI-50 (labial commissure - zygion, same side), PAI-154 (diameter of the iris), PAI-160 (lateral iris - pupil), and PAI-171 (medial iris - pupil, same side). Three of these four variables were identified as non-normal in all categories by the Shapiro-Wilk tests presented above, only PAI-50 was not. As an example, the distributions of PAI-160 are presented in Figure~\ref{fig:PAI_160}. It is interesting to observe that the inclusion of these variables in the proposed machine learning model does not introduce any difficulties, as would occur in a standard regression model where large correlations and/or lack of variation would cause instability in the model estimates. The model simply will not use variables which don't provide additional predictive power.

\section{Results}
\label{sec:results}

This section presents all the results obtained using the proposed artificial neural network architecture detailed in Subsection~\ref{subsec:experimental}, including all the \textit{F}$_{1}$ scores and confusion matrices for each proposed test. 

\subsection{\textit{F$_{1}$} results}

\begin{figure}[!htpb]
\centering
\includegraphics[width=255px]{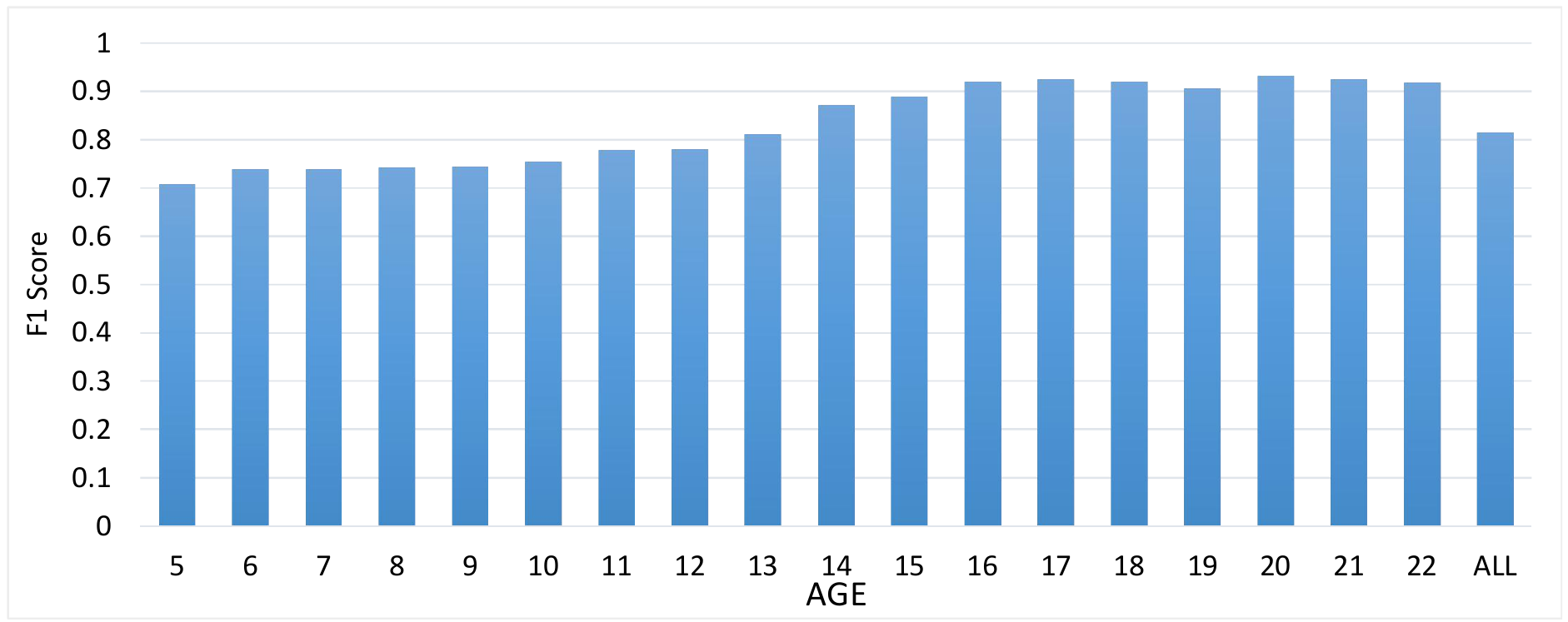}
\caption{Group A: The \textit{F}$_{1}$ score results for sex estimation by each age separately and in the last column the sex estimation test using the whole database.}
\label{fig:f1_sex}
\end{figure}

For \textbf{Group A}, the evaluated tests describe how the PAI indexes can classify the sex information. Figure~\ref{fig:f1_sex} presents the \textit{F}$_{1}$ score for sex estimation for each age. The last column (all) was obtained using all the ages in the classification process in order to evaluate the sex classification baseline test or ``global" test. We can compare the mean of the results for each age to the ``global" test, obtaining similar \textit{F}$_{1}$ scores: $0.83$ and $0.81$ respectively.

\begin{figure}[!htpb]
\centering
\includegraphics[width=255px]{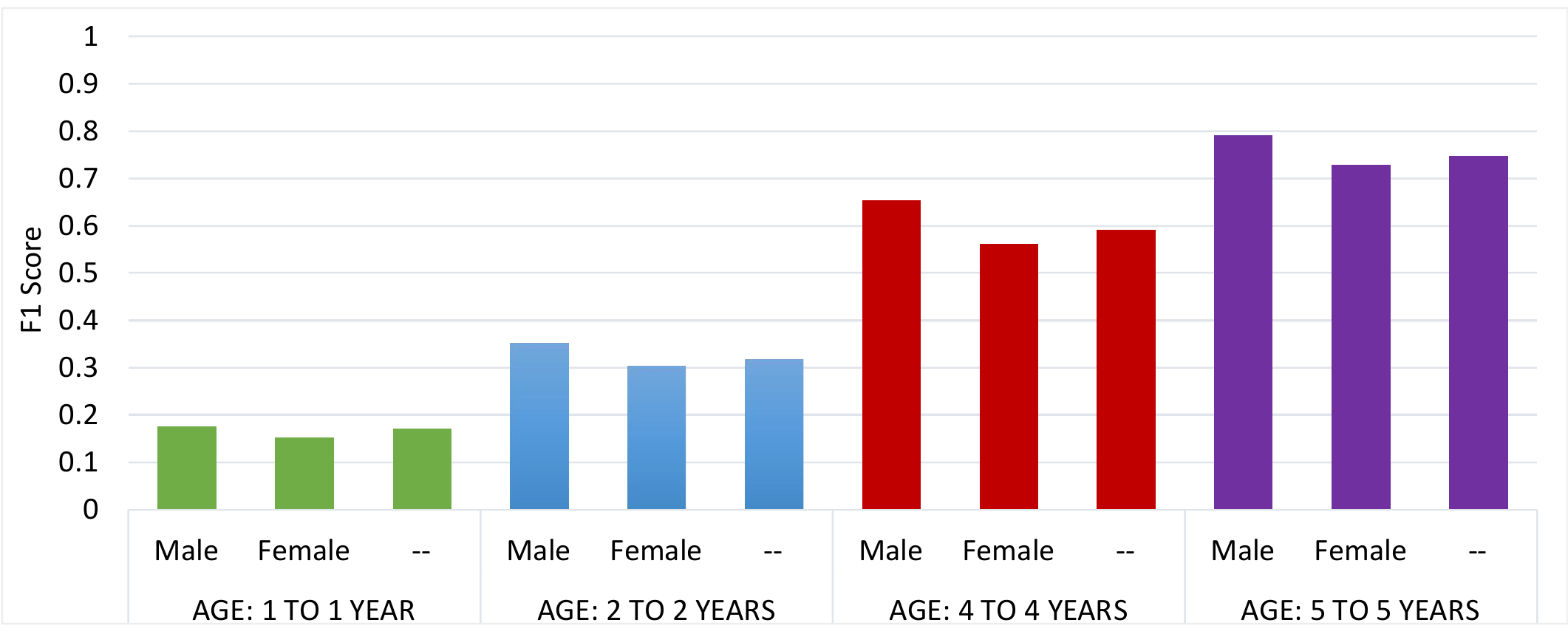}
\caption{Group B: The \textit{F}$_{1}$ score results for age estimation in four scenarios separated by male, female and without sex information.}
\label{fig:f1_age}
\end{figure}

The tests from \textbf{Group B} use the PAI indexes for age classification. Figure~\ref{fig:f1_age} presents the \textit{F}$_{1}$ scores for age estimation divided in four groups: age intervals of 1 year (green), age intervals of 2 years (blue), age intervals of 4 years (red) and age intervals of 5 years (purple). Each group test was divided into three subtests: the first one is the age estimation on male individuals. The second one just female individuals and the last one using both and avoiding the sex information.

\begin{figure}[!htpb]
\centering
\includegraphics[width=255px]{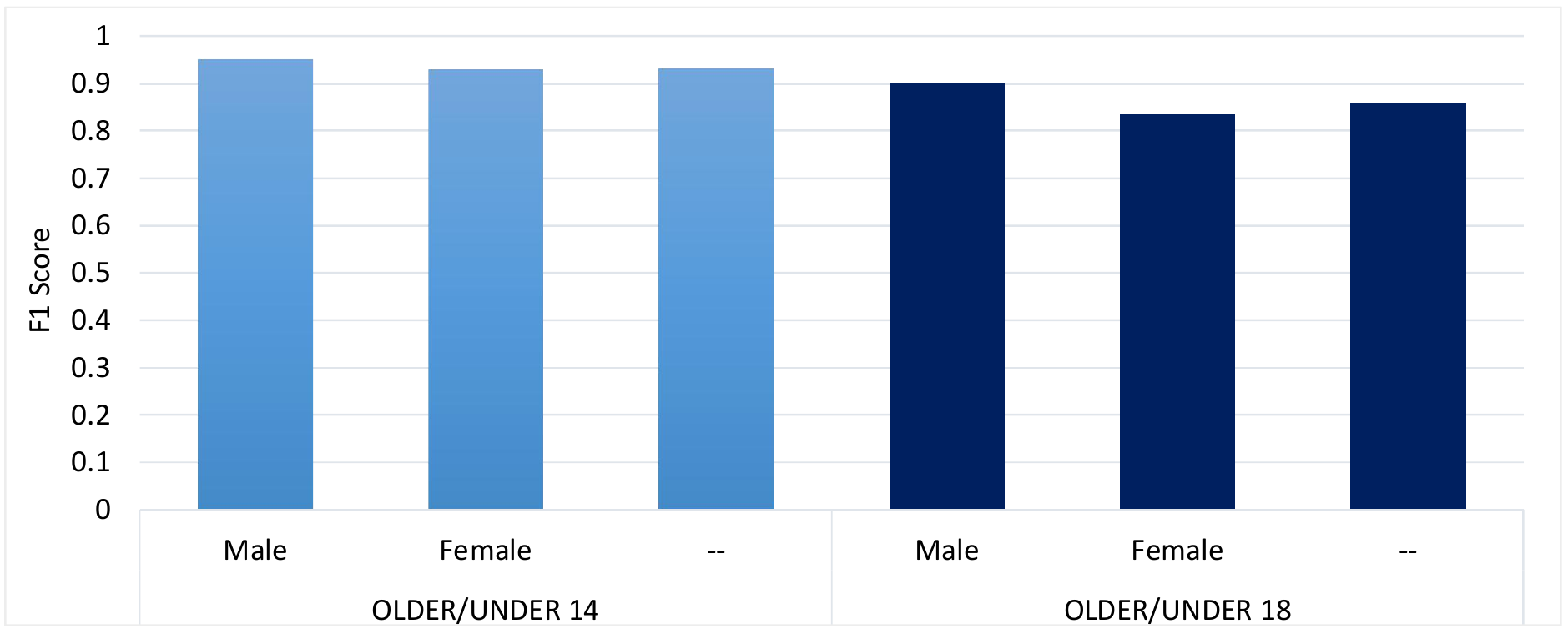}
\caption{Group C: The \textit{F}$_{1}$ score results for age group estimation in two scenarios: threshold of 14 and 18 years old, and both separated by male, female and without sex information.}
\label{fig:f1_agegroups}
\end{figure}

The \textbf{Group C} tests are focused on evaluating if the PAI indexes can classify the age group information correctly. Figure~\ref{fig:f1_agegroups} presents the \textit{F}$_{1}$ scores for age group estimation divided into two tests: older/under 14 years old and older/under 18 years old. Each group test was divided into three subtests, as were the \textbf{Group B} tests: male individuals; female individuals; and all individuals avoiding the sex information.

\subsection{Confusion Matrices}

For each group test, we used confusion matrices to provide a detailed report, including all achieved results for age and sex classification. The matrices demonstrate the classifiers' accuracy by comparing the ``predicted labels" to the ``true labels".

\begin{figure*}[!htpb]
  \centering
  \subfloat[Age 5]  {\label{fig:mf_age_1}  \includegraphics[width=30mm]{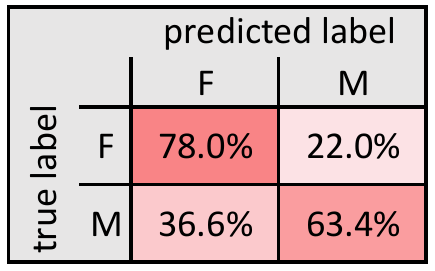}}
  \subfloat[Age 6]  {\label{fig:mf_age_2}  \includegraphics[width=30mm]{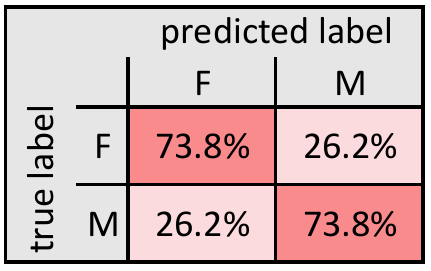}}
  \subfloat[Age 7]  {\label{fig:mf_age_3}  \includegraphics[width=30mm]{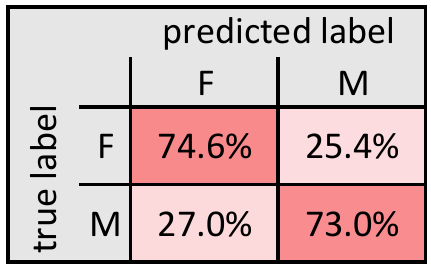}}
  \subfloat[Age 8]  {\label{fig:mf_age_4}  \includegraphics[width=30mm]{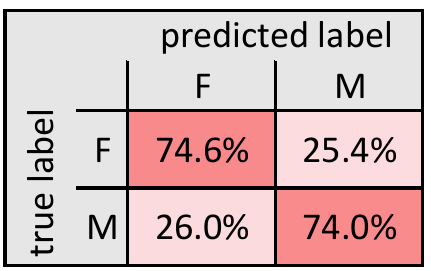}}
  \\
  \subfloat[Age 9]  {\label{fig:mf_age_9}  \includegraphics[width=30mm]{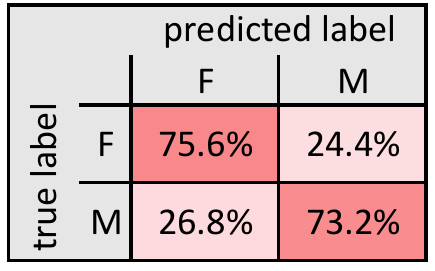}}
  \subfloat[Age 10] {\label{fig:mf_age_10} \includegraphics[width=30mm]{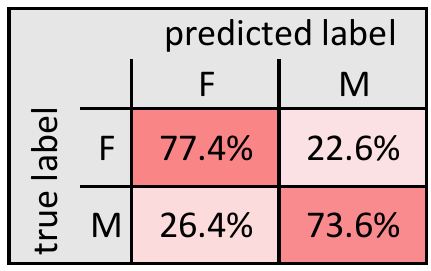}}
  \subfloat[Age 11] {\label{fig:mf_age_11} \includegraphics[width=30mm]{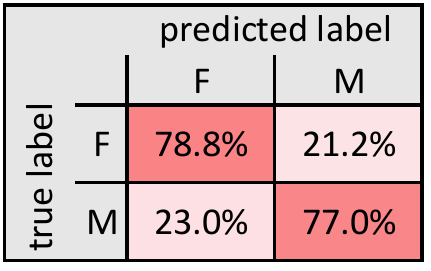}}
  \subfloat[Age 12] {\label{fig:mf_age_12} \includegraphics[width=30mm]{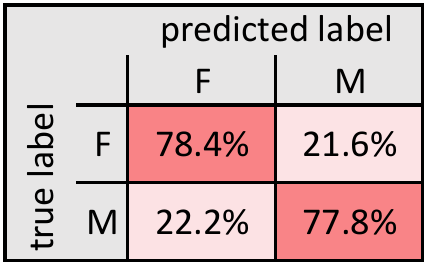}}
  \\
  \subfloat[Age 9]  {\label{fig:mf_age_13} \includegraphics[width=30mm]{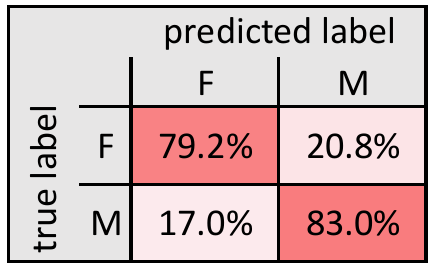}}
  \subfloat[Age 10] {\label{fig:mf_age_14} \includegraphics[width=30mm]{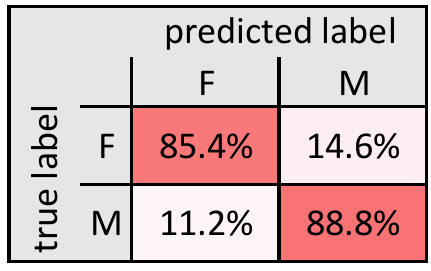}}
  \subfloat[Age 11] {\label{fig:mf_age_15} \includegraphics[width=30mm]{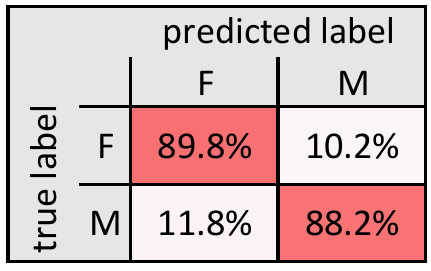}}
  \subfloat[Age 12] {\label{fig:mf_age_16} \includegraphics[width=30mm]{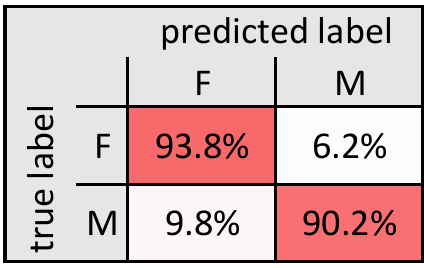}}
  \\
  \subfloat[Age 17] {\label{fig:mf_age_17} \includegraphics[width=30mm]{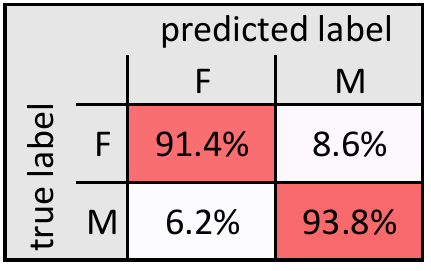}}
  \subfloat[Age 18] {\label{fig:mf_age_18} \includegraphics[width=30mm]{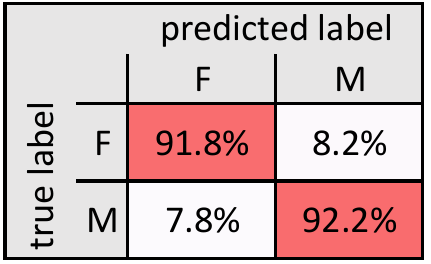}}
  \subfloat[Age 19] {\label{fig:mf_age_19} \includegraphics[width=30mm]{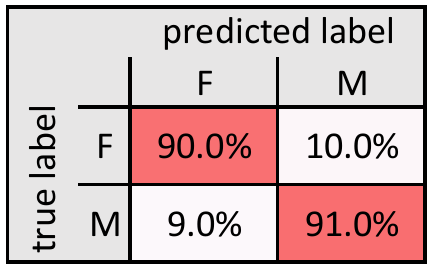}}
  \subfloat[Age 20] {\label{fig:mf_age_20} \includegraphics[width=30mm]{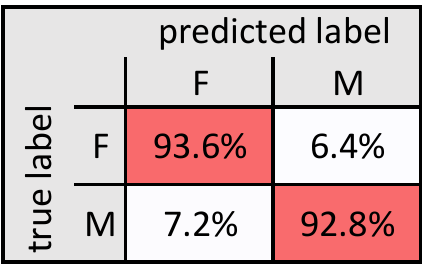}}
  \\
  \subfloat[Age 21] {\label{fig:mf_age_21} \includegraphics[width=30mm]{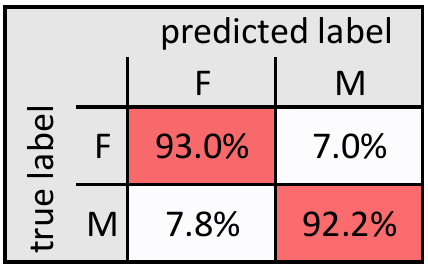}}
  \subfloat[Age 22] {\label{fig:mf_age_22} \includegraphics[width=30mm]{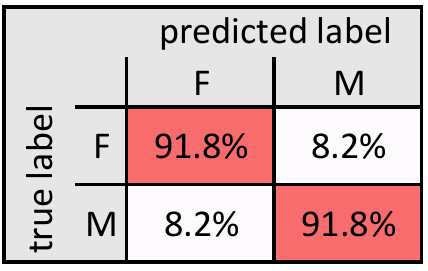}}
  \subfloat[Average]{\label{fig:mf_age_avg}\includegraphics[width=30mm]{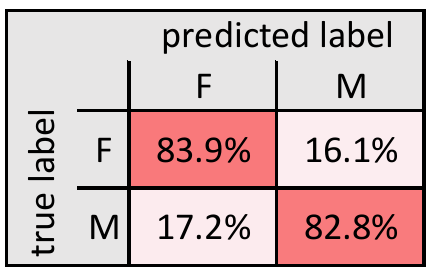}}
  \subfloat[All]    {\label{fig:mf_age_all}\includegraphics[width=30mm]{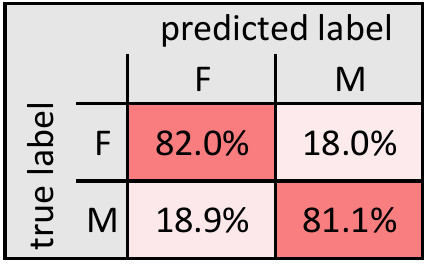}}
  \caption{Matrix confusion for each age group, average and all.}
  \label{fig:mf_age}
\end{figure*}

Figure~\ref{fig:mf_age} presents the confusion matrices for the \textbf{Group A} tests. The figures from Figure~\ref{fig:mf_age}~\subref{fig:mf_age_1} to Figure~\ref{fig:mf_age}~\subref{fig:mf_age_22} present the confusion matrices for each age group separately, from 5 to 22 years old respectively. Figure~\ref{fig:mf_age}~\subref{fig:mf_age_avg} presents the average results of all age groups, meanwhile the Figure~\ref{fig:mf_age}~\subref{fig:mf_age_all} presents the results using all data avoiding the age information.

\begin{figure*}[!htpb]
\centering
\includegraphics[width=1\textwidth]{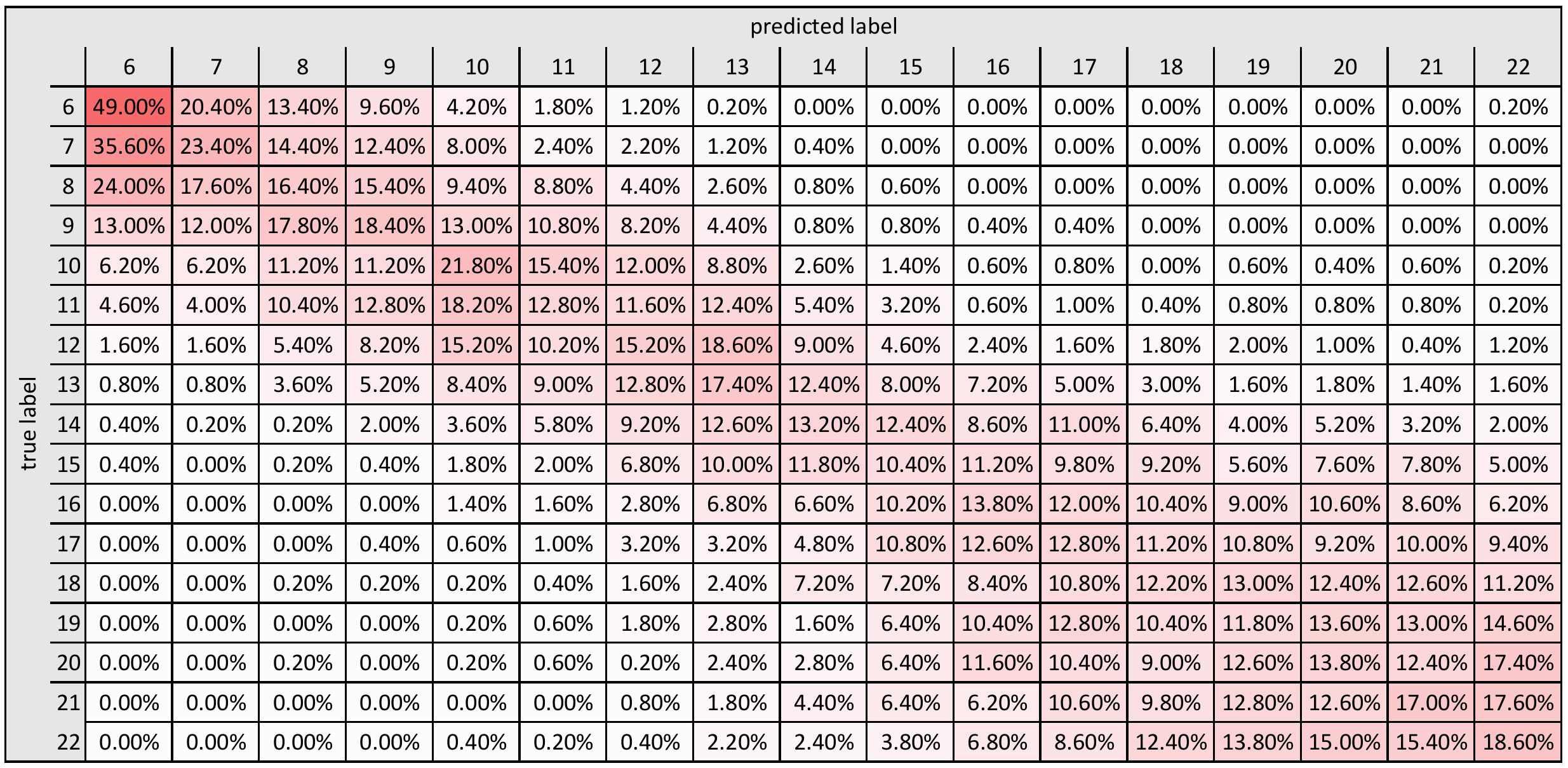}
\caption{Confusion matrix: age estimation at age intervals of 1 year without sex information.}
\label{fig:age_1_1}
\end{figure*}

\begin{figure*}[!htpb]
\centering
\includegraphics[width=1\textwidth]{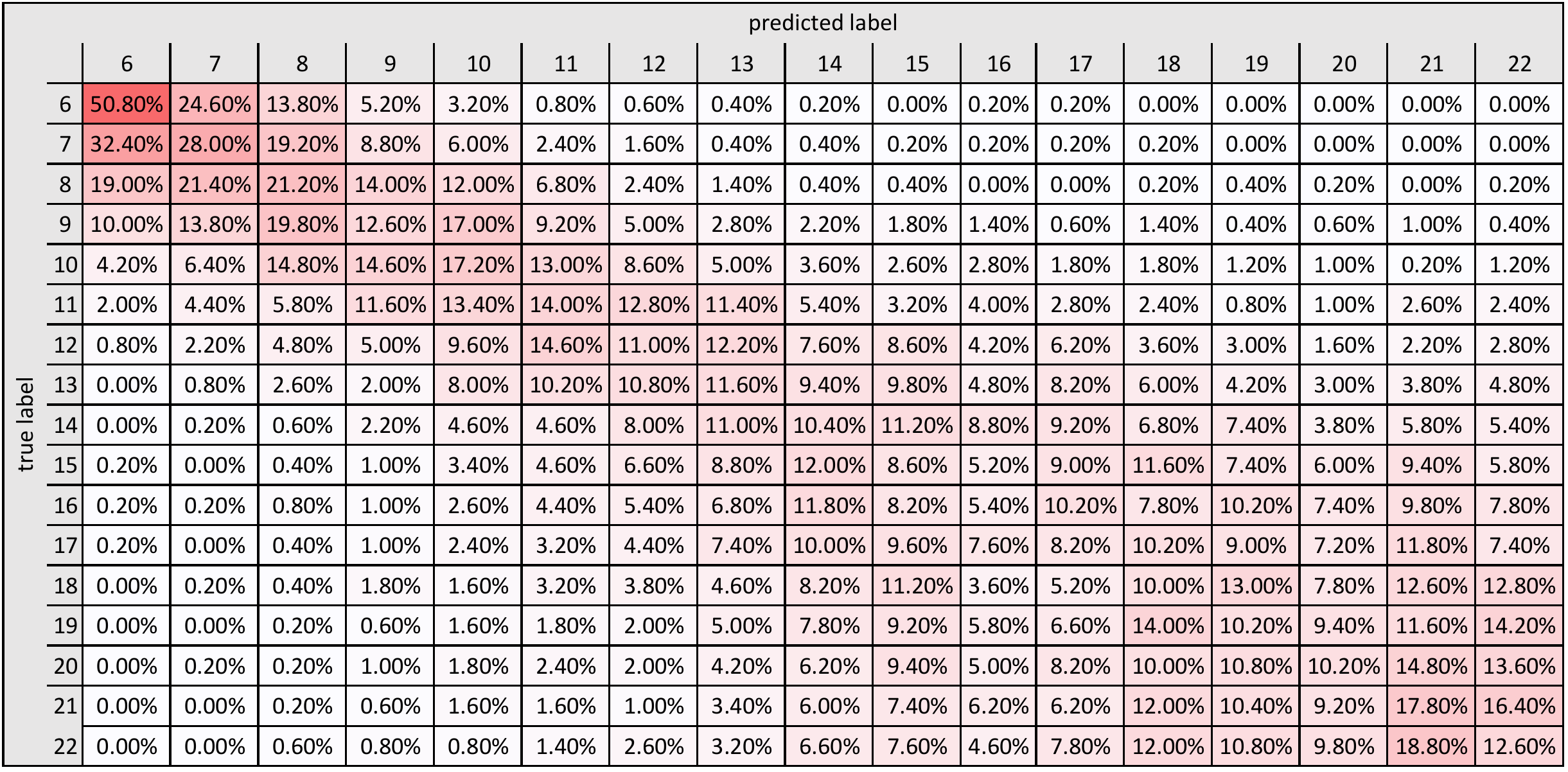}
\caption{Confusion matrix: age estimation at age intervals of 1 year for female sex.}
\label{fig:age_1_1_F}
\end{figure*}

\begin{figure*}[!htpb]
\centering
\includegraphics[width=1\textwidth]{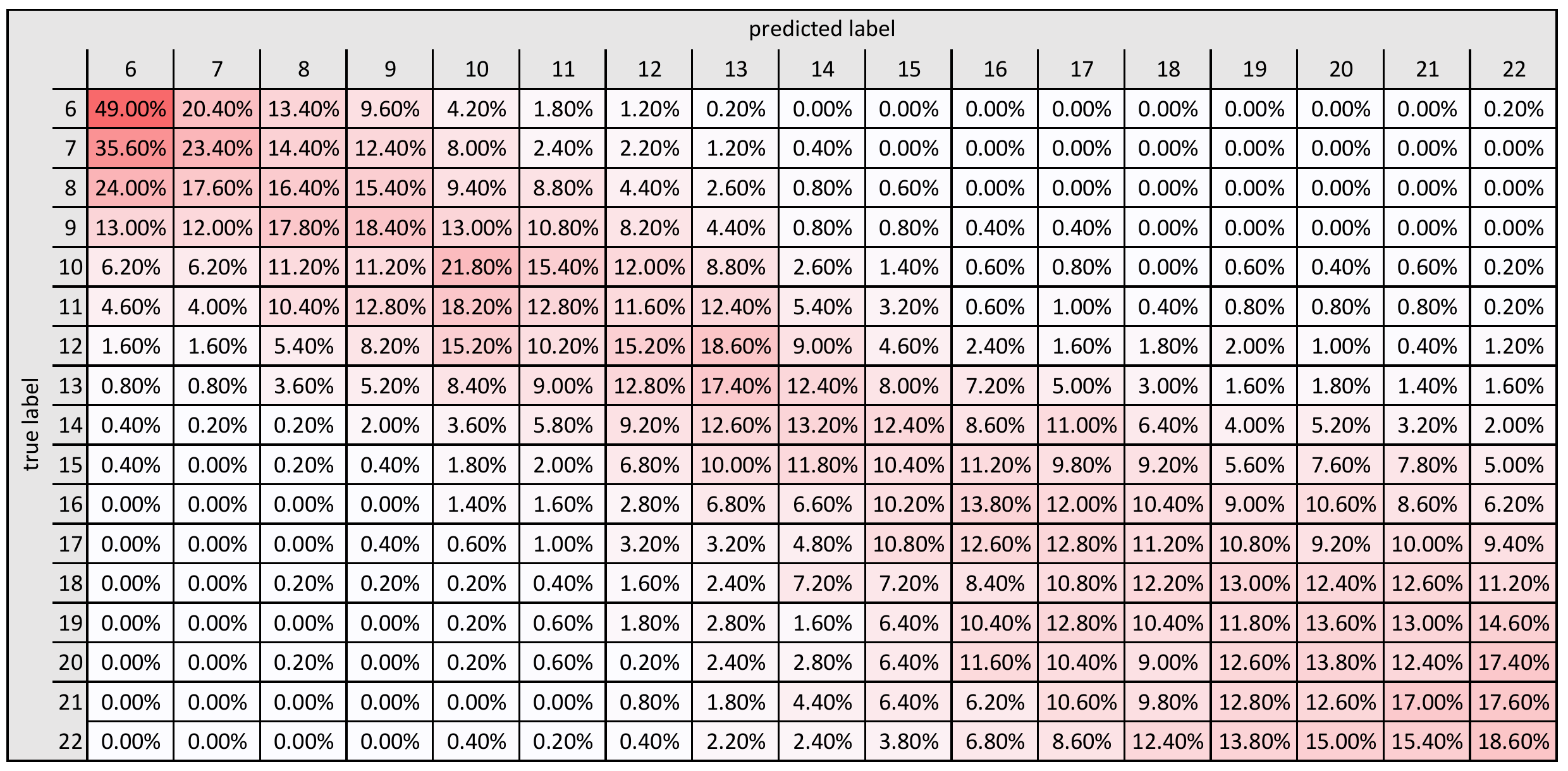}
\caption{Confusion matrix: age estimation at age intervals of 1 year for male sex.}
\label{fig:age_1_1_M}
\end{figure*}

Figures~\ref{fig:age_1_1},~\ref{fig:age_1_1_F} and~\ref{fig:age_1_1_M} present the confusion matrices for tests in Group B using 1 year for age intervals. Figure~\ref{fig:age_1_1} presents the results for age estimation avoiding the sex information. Figure~\ref{fig:age_1_1_F} presents the results for age estimation for females. Figure~\ref{fig:age_1_1_M} presents the results for age estimation for males.

\begin{figure*}[!htpb]
\centering
\includegraphics[width=300px]{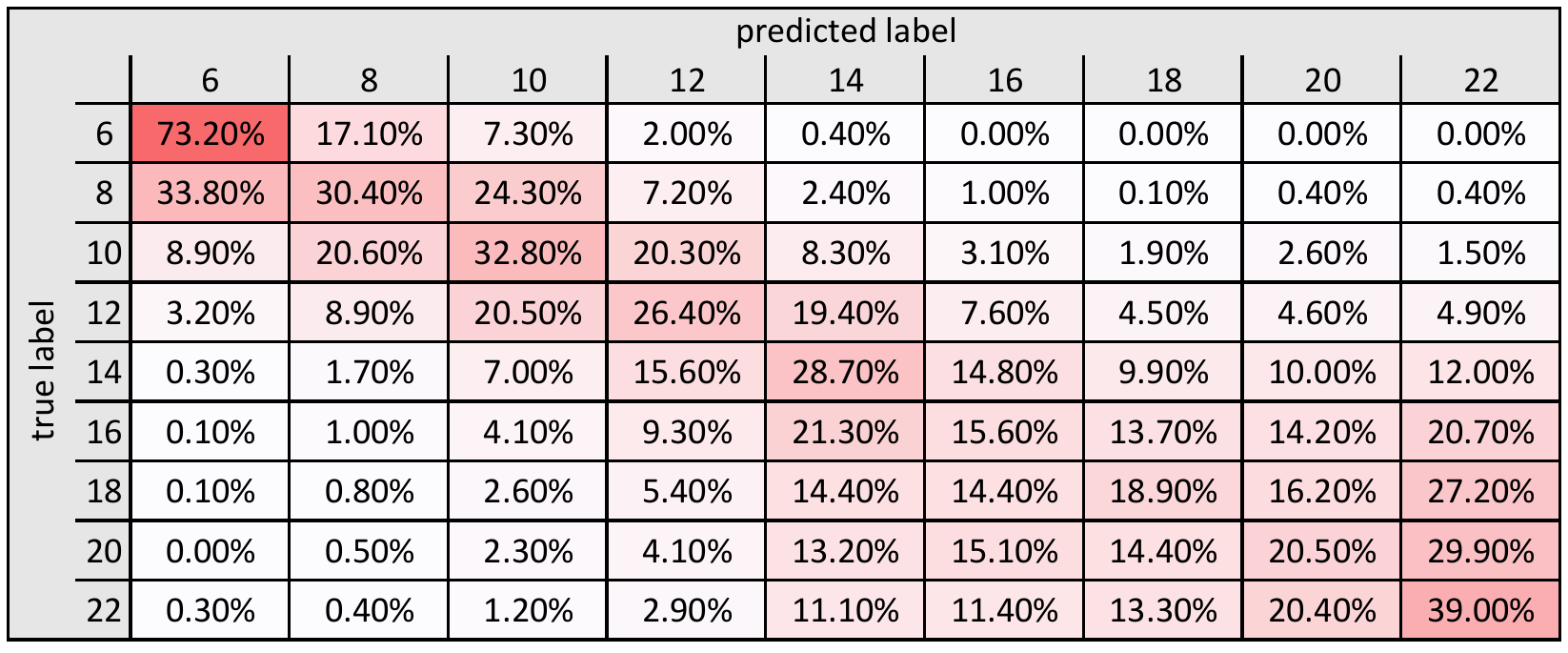}
\caption{Confusion matrix: age estimation at age intervals of 2 years without sex information.}
\label{fig:age_2_2}
\end{figure*}

\begin{figure*}[!htpb]
\centering
\includegraphics[width=300px]{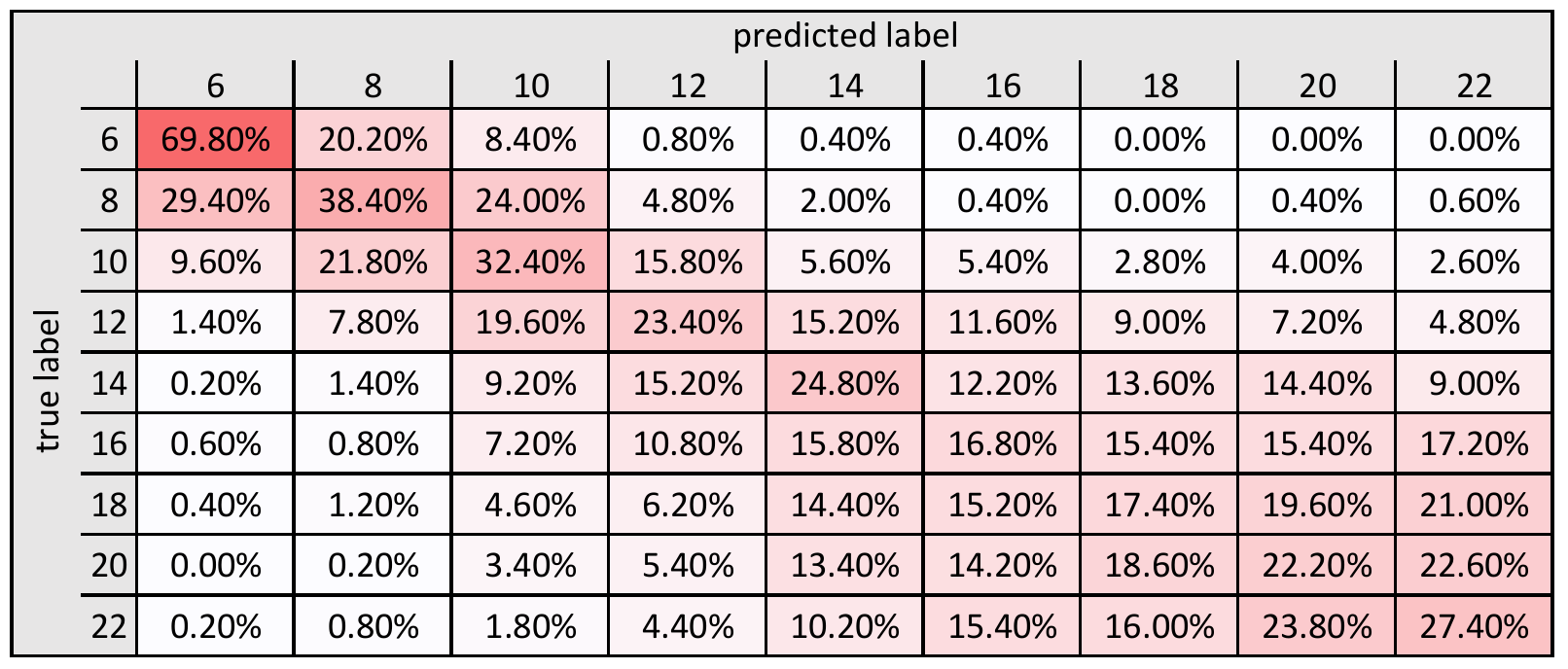}
\caption{Confusion matrix: age estimation at age intervals of 2 years for female sex.}
\label{fig:age_2_2_F}
\end{figure*}

\begin{figure*}[!htpb]
\centering
\includegraphics[width=300px]{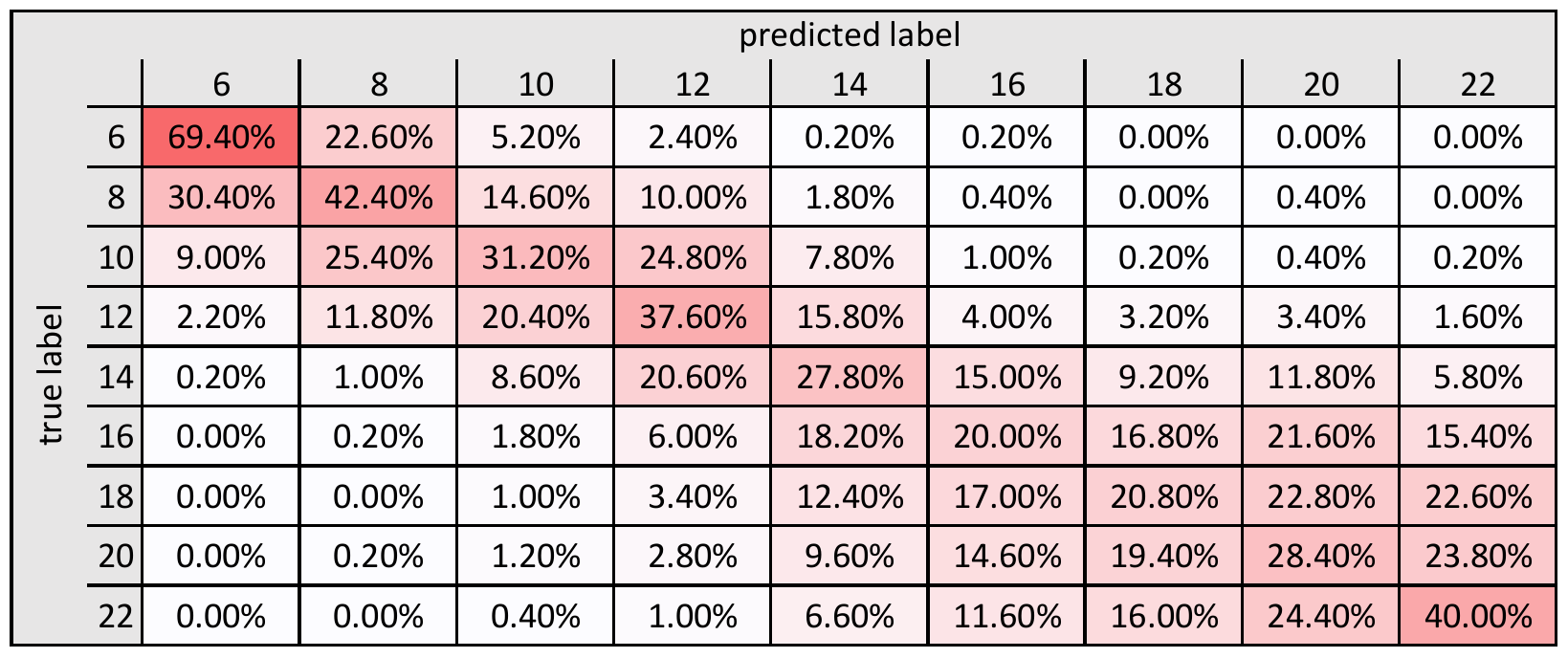}
\caption{Confusion matrix: age estimation at age intervals of 2 years for male sex.}
\label{fig:age_2_2_M}
\end{figure*}

Figures~\ref{fig:age_2_2},~\ref{fig:age_2_2_F} and~\ref{fig:age_2_2_M} present the confusion matrices with the detailed results for Group B: age intervals of 2 years. Figure~\ref{fig:age_2_2} presents the detailed results for age estimation avoiding the sex information. Figure~\ref{fig:age_2_2_F} and Figure~\ref{fig:age_2_2_M} present the detailed results for age estimation for female and male individuals, respectively.



\begin{figure*}[!htpb]
  \centering
  \subfloat[Without sex information.]{\label{fig:age_4_4}\includegraphics[width=50mm]{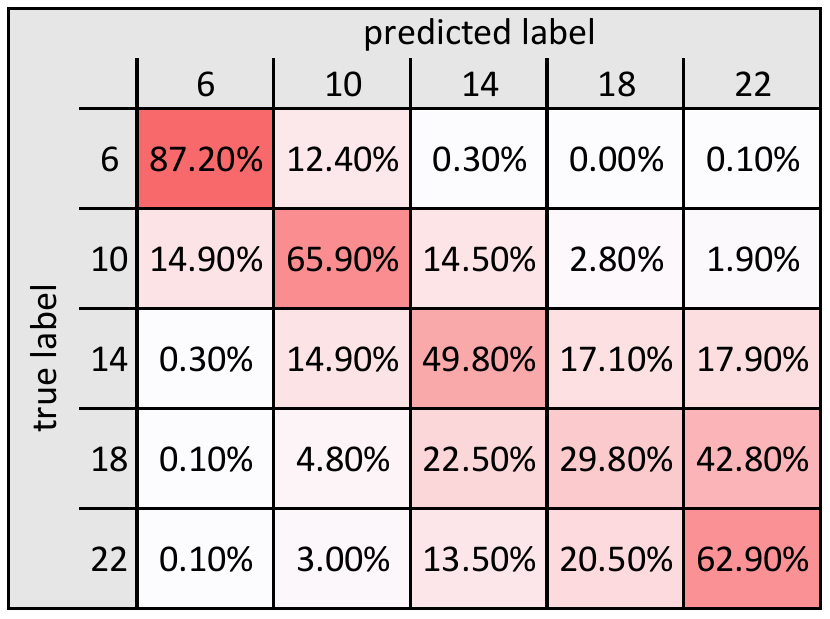}}
  \qquad
  \subfloat[Female sex.]{\label{fig:age_4_4_F}\includegraphics[width=50mm]{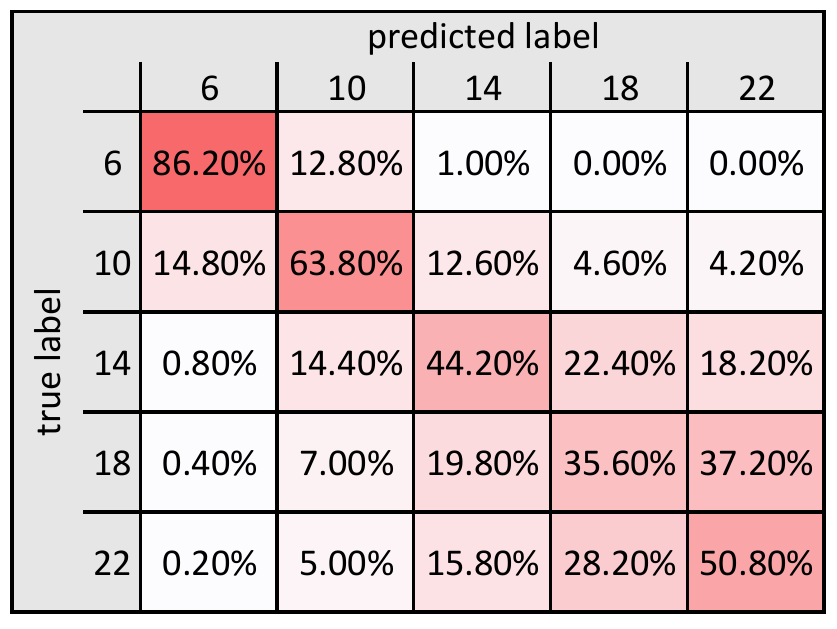}}
  \\
  \subfloat[Male sex.]{\label{fig:age_4_4_M}\includegraphics[width=50mm]{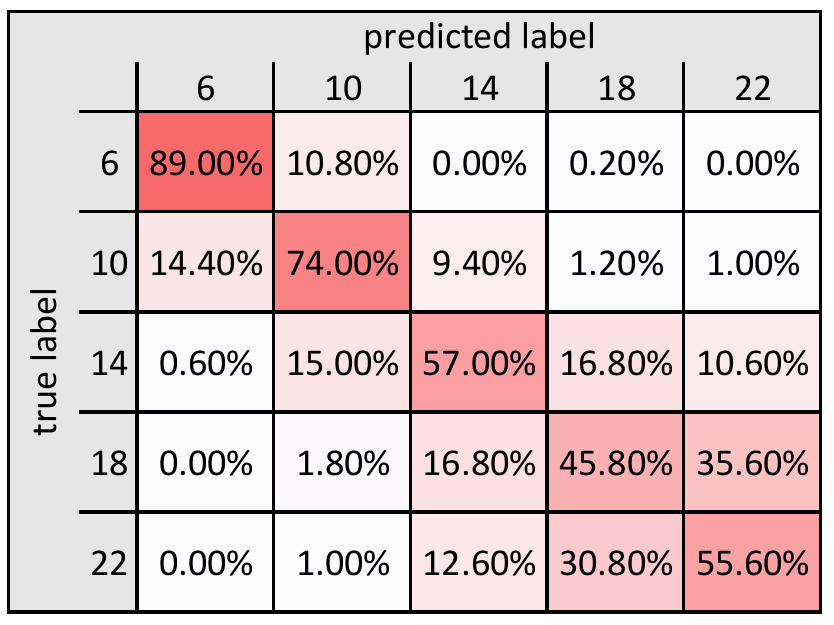}}
  \caption{Confusion matrix: age estimation with age intervals of 4 years.}
  \label{fig:age_4_4_main}
\end{figure*}

Figure~\ref{fig:age_4_4_main} presents the confusion matrices with the detailed results for Group B: age intervals of 4 years. Figure~\ref{fig:age_4_4_main}~\subref{fig:age_4_4} presents the detailed results for age estimation avoiding the sex information. Figure~\ref{fig:age_4_4_main}~\subref{fig:age_4_4_F} presents the detailed results for age estimation on female individuals. Figure~\ref{fig:age_4_4_main}~\subref{fig:age_4_4_M} presents the detailed results for age estimation on male individuals.

\begin{figure*}[!htpb]
  \centering
  \subfloat[Without sex information.]{\label{fig:age_5_5}\includegraphics[width=50mm]{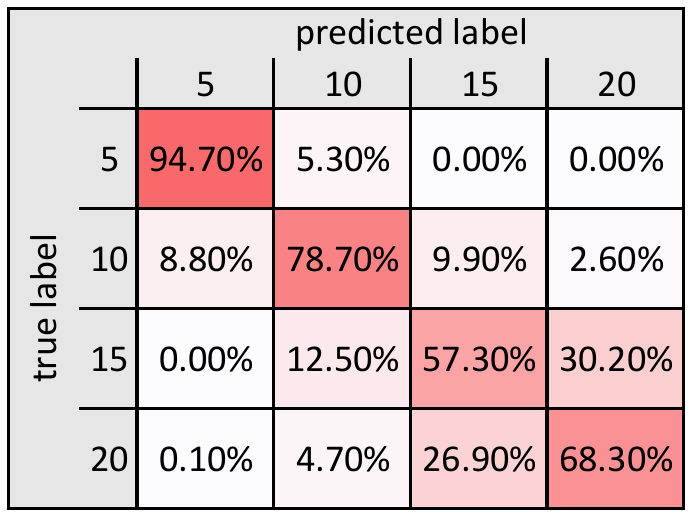}}
  \qquad
  \subfloat[Female sex.]{\label{fig:age_5_5_F}\includegraphics[width=50mm]{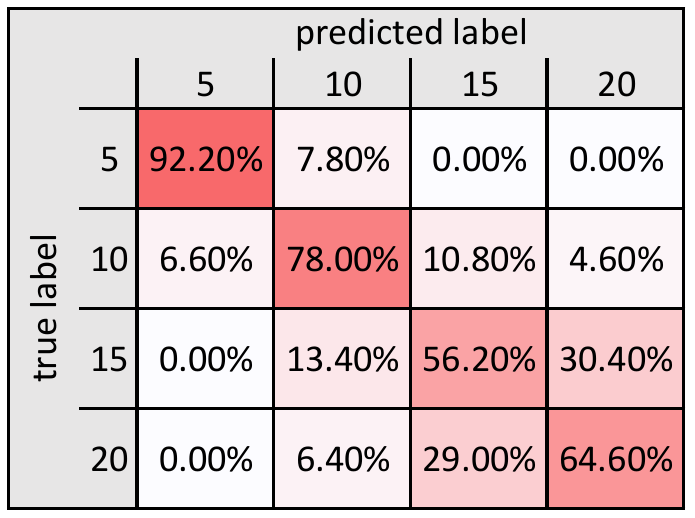}}
  \\
  \subfloat[Male sex.]{\label{fig:age_5_5_M}\includegraphics[width=50mm]{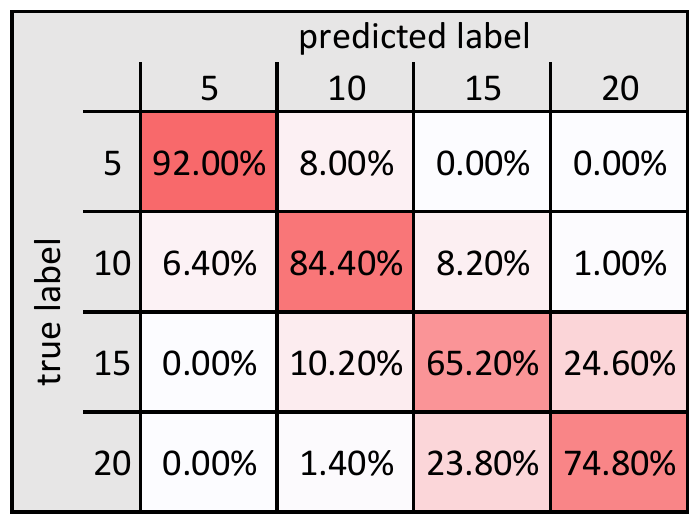}}
  \caption{Confusion matrix: age estimation with age intervals of 5 years.}
  \label{fig:age_5_5_main}
\end{figure*}

Figure~\ref{fig:age_5_5_main} presents the confusion matrices with the detailed results for Group B: age intervals of 5 years. Figure~\ref{fig:age_5_5_main}~\subref{fig:age_5_5} presents the detailed results for age estimation avoiding the sex information. Figure~\ref{fig:age_5_5_main}~\subref{fig:age_5_5_F} presents the detailed results for age estimation on female individuals. Figure~\ref{fig:age_5_5_main}~\subref{fig:age_5_5_M} presents the detailed results for age estimation on male individuals.

\begin{figure*}[!htpb]
  \centering
  \subfloat[Without sex information.]{\label{fig:age_group_14}\includegraphics[width=35mm]{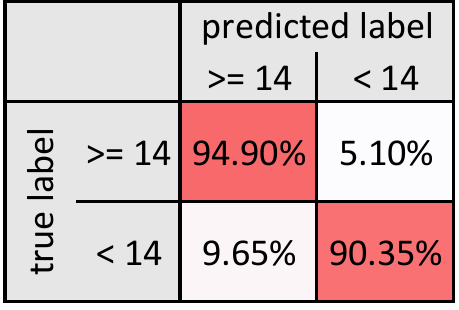}}
  \qquad
  \subfloat[Female sex.]{\label{fig:age_group_14_F}\includegraphics[width=35mm]{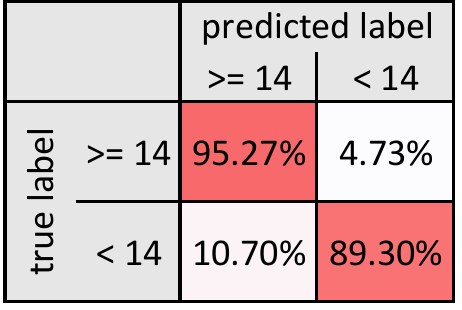}}
  \qquad
  \subfloat[Male sex.]{\label{fig:age_group_14_M}\includegraphics[width=35mm]{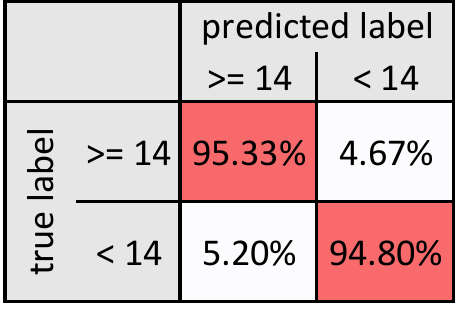}}
  \caption{Age group estimation: threshold of 14 years.}
  \label{fig:AgeGroup14}
\end{figure*}

Figure~\ref{fig:AgeGroup14} presents the confusion matrices for Group C: older/under 14 years old. Figure~\ref{fig:AgeGroup14}~\subref{fig:age_group_14} presents the detailed results for age group estimation avoiding the sex information. Figure~\ref{fig:AgeGroup14}~\subref{fig:age_group_14_F} presents the detailed results for age group estimation on female individuals.  Figure~\ref{fig:AgeGroup14}~\subref{fig:age_group_14_M} presents the detailed results for age group estimation on male individuals.

\begin{figure*}[!htpb]
  \centering
  \subfloat[Without sex information.]{\label{fig:age_group_18}\includegraphics[width=35mm]{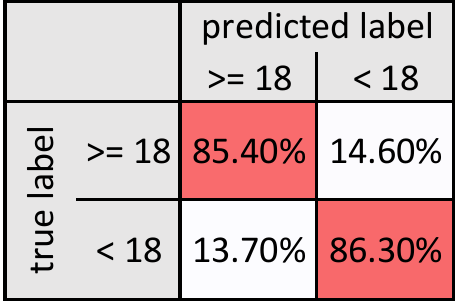}}
  \qquad
  \subfloat[Female sex.]{\label{fig:age_group_18_F}\includegraphics[width=35mm]{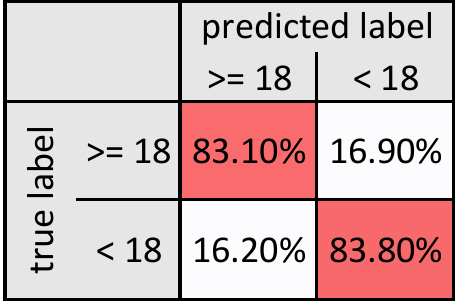}}
  \qquad
  \subfloat[Male sex.]{\label{fig:age_group_18_M}\includegraphics[width=35mm]{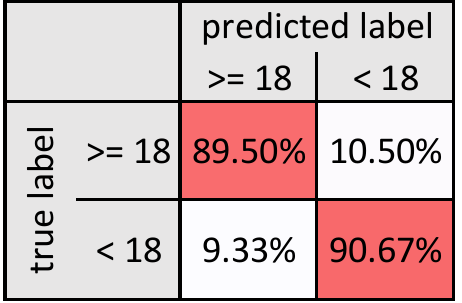}}
  \caption{Age group estimation: threshold of 18 years.}
  \label{fig:AgeGroup18}
\end{figure*}

Figure~\ref{fig:AgeGroup18} presents the confusion matrices with the detailed results for Group C: older/under 18 years old. Figure~\ref{fig:AgeGroup18}~\subref{fig:age_group_18} presents the detailed results for age group estimation avoiding the sex information. Figure~\ref{fig:AgeGroup18}~\subref{fig:age_group_18_F} presents the detailed results for age group estimation on female individuals. Figure~\ref{fig:AgeGroup18}~\subref{fig:age_group_18_M} presents the detailed results for age group estimation on male individuals.

\section{Discussion}
\label{sec:conclusion}

Estimating the age of victims and suspects of crimes is an essential procedure in the forensic casuistic of human identification~\cite{schmeling2001age,schmeling2016forensic,silva2013evidencia,machado2017age}. Age estimation becomes even more important when legal age thresholds are determined by the Court~\cite{deitos2015age,santiago2018accuracy}. Currently, the ages of 14 and 18 years represent the legal age threshold of sexual consent and majority in the judicial system of several countries worldwide~\cite{franco2013applicability,graupner2000sexual}. Designing scientific tools that allow the investigation of age with accuracy and evidence-based standards must be continuously encouraged to promote optimal forensic practices. Justified by the need for improving facial age estimation through photo-anthropometric analysis and founded on the hypothesis that the photo-anthropometric analyses of the human face can distinguish individuals younger and older than 14 and 18 years, this study aimed to propose and test an automatic solution to distinguish (male and female) individuals younger and older than 14 and 18 using photo-anthropometric analyses of the human face.

Population-specific studies on the use of photo-anthropometry of the face for age estimation are scarce in the scientific literature – especially those with large and standardized samples~\cite{palhares2017,flores2018comparative,gonzales2018photoanthropometry}. In this study, a large sample of Brazilian participants was collected (n=18000) and organized in a detailed, standardized data set available only for academic purposes. Sample standardization was accomplished by selecting male (n=500) and female (n=500) participants equally distributed (n=1000) in age intervals of one year (from 5 to 22 years). Moreover, all the digital photographs of the participants were taken with the same equipment and followed protocols previously reported in the scientific literature~\cite{palhares2017} and by the International Civil Aviation Organization (ICAO). In the scope of sampling strategy, an additional gain was obtained by choosing Brazilian participants, which are known for their multiracial formation~\cite{souza2016science} and phenotypic diversity~\cite{edmonds2007triumphant}. This characteristic makes the study outcomes novel and eventually more reproducible in other populations.

In relation to the study outcomes, inferences about age were first investigated in association with sex. More specifically, the morphological information retrieved from the human face was tested based on its performance to classify males and females in age categories (5-22). A separate statistical test was evaluated for each of the 18 age groups. The outcomes endorsed the scientific literature by revealing difficulties for sexual dimorphism in young participants ($<$12 years). In particular, the mean \textit{F}$_{1}$ scores in young age groups ranged between $0.7$ and $0.78$. On the other hand, for participants aged from 13 to 22 years the mean \textit{F}$_{1}$ scores increased from $0.81$ to $0.91$, showing evident improvement for classifying males and females. According to the scientific literature, the difference between young and old participants is explained by the lack of mature secondary sexual features depicted in facial photographs of children~\cite{kloess2017challenges}. 

Recently, Kloess in~\cite{kloess2017challenges} investigated the challenges in classifying child sexual abuse images. In their study, inferences about age (minor or not) were more easily and accurately given in images of babies and toddlers, while difficulties increased when the age of interest approached young adolescence. Interestingly, the larger size of the eyes in comparison with the other facial features and the presence of (milk) teeth and interdental gaps emerged as potential indicators of youth, while the use of make-up was a confounding factor~\cite{kloess2017challenges}. In the present study, the diameter of the iris (iris ratio)~\cite{palhares2017} was used as fixed reference to calculate morphological ratios from the human face – which enables a quantification of the qualitative information provided by the previous authors. Additionally, photographs of participants depicting unnatural facial expressions (e.g. smiling) or using make-up were part of the exclusion criteria in the present study. Corroborated by the scientific literature~\cite{kloess2017challenges}, this methodological set up promoted a reduction of age classification bias as function of sex. In this context, sexual dimorphism, regardless of age, was performed as a quality-control procedure to eliminate the influence of age over the classification performance. The mean \textit{F}$_{1}$ score reached $0.81$, indicating that the classification tool was able to properly distinguish most of the males and females of the sample if they were combined in a single group. In practice, the use of age-specific classification tools is recommended (whenever applicable) to best-fit the needs of each case – especially if the case involves age interests between 13 and 22 years, such as the age of sexual consent~\cite{graupner2000sexual} and the age of legal majority~\cite{franco2013applicability}.

In a second phase, the present study engaged in a deeper investigation based on age. This phase was justified by the uncertainty regarding the chronological age of victims and suspects of crimes that are commonly observed during the routine of forensic services. The methodological set up at this phase clustered together not only participants in age intervals of one year, but also in larger age intervals (e.g. within two, four and five years). Within each group, the performance was better for classifying the age of males. Between groups, the mean \textit{F}$_{1}$ scores were progressively higher with the increase in age interval size. Consequently, the best age estimates were found in the group with age interval of five years (mean \textit{F}$_{1}$ score: 0.74 combining males and females). Clearly, this outcome shows that the classification process becomes more difficult by refining the sample based on age. A similar approach was recently used by Machado in~\cite{palhares2017}. The authors performed a photo-anthropometric analysis to investigate the allometric growth of the human face by grouping together individuals within age intervals of four years. Despite the evident contributions towards the analysis of facial alterations over the time, the methodological set up proposed by the authors was limited compared to the present study. The advantages highlighted in the present setup include not only the sample stratification in groups of four different age intervals, but also the collection and quantification of much more morphological information from the human face. While the authors in~\cite{palhares2017} mapped human facial growth with ten measurements calculated from ratios based on the diameter of the iris, the present study outcomes were founded on 208 measures calculated with the same rationale. In practice, the improved methodological setup proposed in the present study induces more reliable and accurate age estimations.

The third and final phase addressed in this study focused on testing the classification system to distinguish participants that were younger or older than specific legal age thresholds of interest. This set up was justified to specifically meet the needs of justice when it comes to answer legal requests regarding the ages of 14 – related to sexual consent, and 18 – related to legal majority. The tests that were carried out in this phase showed, again, better classification of males, both in relation to the threshold of 14 and 18 years. When males and females were combined, the mean \textit{F}$_{1}$ scores reached 0.93 and 0.85 for the ages of 14 and 18 years, respectively. Satisfactory outcomes were also recently observed by Borges in~\cite{BORGES20189} with a similar approach. The authors obtained accuracy (Area Under the Curve) estimates from Receiver Operating Characteristic (ROC) curve analyses that reached 0.96 and 0.90 for the ages of 14 and 18, respectively. Differences between studies include the larger sample size in the current investigation (n=18000 in face of 1000 used by the authors~\cite{BORGES20189}) and, again, the larger number of measurements from the human face (n=208 compared to the 40 measurements used by the authors~\cite{BORGES20189}). In forensic practice, the performance of the classification systems used in this study strengthens and supports its use for distinguishing victims and suspects of crimes aged below or above 14 and 18.

The automatic solution developed to classify individuals based on age and sex using morphological information retrieved from photo-anthropometric analyses of the human face reached optimal outcomes. More accurate age estimates were found in subjects aged between 13 and 22 years; estimates for the age of males were better than for females; and classifications based on legal age thresholds of sexual consent and majority were feasible and promising.

The proposed methodological setup presents multiple advantages compared to the available scientific literature. However, translating it to practice requires careful implementations and follow-up work to include updates in scientific evidence. Future studies in the field should test the reproducibility of this methodological setup and inherent outcomes in different populations. Investigations based on other legal age thresholds of interest are also encouraged to best-fit the judicial systems of different countries. Advances in the methodological setup for further improvements should include longitudinal sampling and three-dimensional imaging. Another approach in the computer vision field is to evaluate a classifier using deep learning techniques for age and sex estimation~\cite{cao2018vggface2,xing2017diagnosing}, combining photos with photo-anthropometric indexes creating a cross-domain classifier. This allows the evaluation of possible improvements achieved by the inclusion of PAIs when compared to using only images as input.


%

\section*{Acknowledgment}
The authors would like to acknowledge the team of Federal Police of Brazil, specially the forensic experts of National Institute of Criminalistic. This work was conducted with financial support from Coordination for the Improvement of Higher Education Personnel (CAPES) and Federal Police of Brazil (Grant Number: 001 Pro-Forenses 25/2014 CAPES).


\ifCLASSOPTIONcaptionsoff
  \newpage
\fi



%

\bibliographystyle{IEEEtran}
\bibliography{ref}

\begin{thebibliography}{10}
\providecommand{\url}[1]{#1}
\csname url@samestyle\endcsname
\providecommand{\newblock}{\relax}
\providecommand{\bibinfo}[2]{#2}
\providecommand{\BIBentrySTDinterwordspacing}{\spaceskip=0pt\relax}
\providecommand{\BIBentryALTinterwordstretchfactor}{4}
\providecommand{\BIBentryALTinterwordspacing}{\spaceskip=\fontdimen2\font plus
\BIBentryALTinterwordstretchfactor\fontdimen3\font minus
  \fontdimen4\font\relax}
\providecommand{\BIBforeignlanguage}[2]{{%
\expandafter\ifx\csname l@#1\endcsname\relax
\typeout{** WARNING: IEEEtran.bst: No hyphenation pattern has been}%
\typeout{** loaded for the language `#1'. Using the pattern for}%
\typeout{** the default language instead.}%
\else
\language=\csname l@#1\endcsname
\fi
#2}}
\providecommand{\BIBdecl}{\relax}
\BIBdecl

\bibitem{marquez2015overview}
N.~Marquez-Grant, ``An overview of age estimation in forensic anthropology:
  perspectives and practical considerations,'' \emph{Annals of human biology},
  vol.~42, no.~4, pp. 308--322, 2015.

\bibitem{silva2013interrelationship}
R.~Silva, A.~Franco, P.~Dias, A.~Gon{\c{c}}alves, and L.~Paranhos,
  ``Interrelationship between forensic radiology and forensic odontology—a
  case report of identified skeletal remains,'' \emph{Journal of Forensic
  Radiology and Imaging}, vol.~1, no.~4, pp. 201--206, 2013.

\bibitem{adserias2018forensic}
J.~Adserias-Garriga, C.~Thomas, D.~H. Ubelaker, and S.~C. Zapico, ``When
  forensic odontology met biochemistry: Multidisciplinary approach in forensic
  human identification,'' \emph{Archives of oral biology}, vol.~87, pp. 7--14,
  2018.

\bibitem{interpol2018DVI-Lion}
\BIBentryALTinterwordspacing
INTERPOL, ``Interpol disaster victim identification guide,''
  \url{https://www.interpol.int/INTERPOL-expertise/Forensics/DVI-Pages/DVI-guide},
  2018. [Online]. Available: \url{https://www.interpol.int/}
\BIBentrySTDinterwordspacing

\bibitem{cattaneo2012can}
C.~Cattaneo, Z.~Obertov{\'a}, M.~Ratnayake, L.~Marasciuolo, J.~Tutkuviene,
  P.~Poppa, D.~Gibelli, P.~Gabriel, and S.~Ritz-Timme, ``Can facial proportions
  taken from images be of use for ageing in cases of suspected child
  pornography? a pilot study,'' \emph{International journal of legal medicine},
  vol. 126, no.~1, pp. 139--144, 2012.

\bibitem{ratnayake2014juvenile}
M.~Ratnayake, Z.~Obertov{\'a}, M.~Dose, P.~Gabriel, H.~Br{\"o}ker,
  M.~Brauckmann, A.~Barkus, R.~Rizgeliene, J.~Tutkuviene, S.~Ritz-Timme
  \emph{et~al.}, ``The juvenile face as a suitable age indicator in child
  pornography cases: a pilot study on the reliability of automated and visual
  estimation approaches,'' \emph{International journal of legal medicine}, vol.
  128, no.~5, pp. 803--808, 2014.

\bibitem{BORGES20189}
D.~L. Borges, F.~B. Vidal, M.~R. Flores, R.~F. Melani, M.~A. Guimarães, and
  C.~E. Machado, ``Photoanthropometric face iridial proportions for age
  estimation: An investigation using features selected via a joint mutual
  information criterion,'' \emph{Forensic Science International}, vol. 284, pp.
  9 -- 14, 2018.

\bibitem{cattaneo2009difficult}
C.~Cattaneo, S.~Ritz-Timme, P.~Gabriel, D.~Gibelli, E.~Giudici, P.~Poppa,
  D.~Nohrden, S.~Assmann, R.~Schmitt, and M.~Grandi, ``The difficult issue of
  age assessment on pedo-pornographic material,'' \emph{Forensic science
  international}, vol. 183, no.~1, pp. e21--e24, 2009.

\bibitem{palhares2017}
C.~E.~P. Machado, M.~R.~P. Flores, L.~N.~C. Lima, R.~L.~R. Tinoco, A.~Franco,
  A.~C.~B. Bezerra, M.~P. Evison, and M.~A. Guimar{\~{a}}es, ``A new approach
  for the analysis of facial growth and age estimation: Iris ratio,''
  \emph{{PLOS} {ONE}}, vol.~12, no.~7, p. e0180330, 2017.

\bibitem{flores2018comparative}
M.~R. Flores, C.~E. Machado, M.~D. Gallidabino, G.~H. de~Arruda, R.~H.
  da~Silva, F.~B. de~Vidal, and R.~F. Melani, ``Comparative assessment of a
  novel photo-anthropometric landmark-positioning approach for the analysis of
  facial structures on two-dimensional images,'' \emph{Journal of forensic
  sciences}, 2018.

\bibitem{gonzales2018photoanthropometry}
P.~S. Gonzales, C.~E.~P. Machado, and E.~Michel-Crosato, ``Photoanthropometry
  of the face in the young white brazilian population,'' \emph{Brazilian dental
  journal}, vol.~29, no.~6, pp. 619--623, 2018.

\bibitem{zhu2017trends}
G.~Zhu and S.~van~der Aa, ``Trends of age of consent legislation in europe: A
  comparative study of 59 jurisdictions on the european continent,'' \emph{New
  Journal of European Criminal Law}, vol.~8, no.~1, pp. 14--42, 2017.

\bibitem{carpenter2014harm}
B.~Carpenter, E.~O’Brien, S.~Hayes, and J.~Death, ``Harm, responsibility,
  age, and consent,'' \emph{New Criminal Law Review: In International and
  Interdisciplinary Journal}, vol.~17, no.~1, pp. 23--54, 2014.

\bibitem{cericato2016correlating}
G.~O. Cericato, A.~Franco, M.~A.~V. Bittencourt, M.~A.~P. Nunes, and L.~R.
  Paranhos, ``Correlating skeletal and dental developmental stages using
  radiographic parameters,'' \emph{Journal of forensic and legal medicine},
  vol.~42, pp. 13--18, 2016.

\bibitem{machado2018effectiveness}
M.~A. Machado, E.~D. J{\'u}nior, M.~M. Fernandes, I.~F.~P. Lima, G.~O.
  Cericato, A.~Franco, and L.~R. Paranhos, ``Effectiveness of three age
  estimation methods based on dental and skeletal development in a sample of
  young brazilians,'' \emph{Archives of oral biology}, vol.~85, pp. 166--171,
  2018.

\bibitem{ISO19794-5}
{International Organization for Standardization}, ``{ISO/IEC 19794-5:
  Information technology -- Biometric data interchange formats -- Part 5: Face
  image data},'' International Organization for Standardization, Standard, Mar.
  2005.

\bibitem{flores2017manual}
\BIBentryALTinterwordspacing
M.~R. Pinheiro-Flores and C.~E. Palhares-Machado, \emph{Manual of facial
  photoanthropometry: landmarks in frontal view from visual references},
  1st~ed., 2017. [Online]. Available:
  \url{http://facisgroup.org/facial\_landmarks}
\BIBentrySTDinterwordspacing

\bibitem{flores2014master}
M.~R. Pinheiro-Flores, ``Proposta de metodologia de an{\' a}lise
  fotoantropom{\' e}trica para identifica{\c c}{\~a}o humana em imagens faciais
  em norma frontal,'' Master's thesis, Faculdade de Odontologia de Ribeir{\~a}o
  Preto, Universidade de S{\~a}o Paulo, 2014.

\bibitem{PrePrintFSI2019}
\BIBentryALTinterwordspacing
L.~F. Porto, L.~N.~C. Lima, M.~Flores, A.~Valsecchi, O.~Ibanez, C.~E.~M.
  Palhares, and F.~de~Barros~Vidal. (2019, apr) Automatic cephalometric
  landmarks detection on frontal faces: an approach based on supervised
  learning techniques. [Online]. Available:
  \url{https://arxiv.org/abs/1904.10816v1}
\BIBentrySTDinterwordspacing

\bibitem{stephanStandards16}
J.~Caple and C.~Stephan, ``A standardized nomenclature for craniofacial and
  facial anthropometry,'' \emph{International Journal of Legal Medicine}, vol.
  130, no.~3, pp. 863--879, 2016.

\bibitem{brown2004survey}
R.~E. Brown, T.~P. Kelliher, P.~H. Tu, W.~D. Turner, M.~A. Taister, and K.~W.
  Miller, ``A survey of tissue-depth landmarks for facial approximation,''
  \emph{Forensic Sci. Commun}, vol.~6, no.~1, 2004.

\bibitem{phillips2000feret}
P.~J. Phillips, H.~Moon, S.~Rizvi, P.~J. Rauss \emph{et~al.}, ``The feret
  evaluation methodology for face-recognition algorithms,'' \emph{Pattern
  Analysis and Machine Intelligence, IEEE Transactions on}, vol.~22, no.~10,
  pp. 1090--1104, 2000.

\bibitem{intelDev}
\BIBentryALTinterwordspacing
S.~Apeland, ``Intel ai devcloud,'' \url{https://www.intel.ai/devcloud/},
  accessed: 2019-01-22. [Online]. Available:
  \url{https://www.intel.ai/devcloud/}
\BIBentrySTDinterwordspacing

\bibitem{chollet2015keras}
F.~Chollet \emph{et~al.}, ``Keras,'' https://keras.io, 2015.

\bibitem{tensorflow2015-whitepaper}
\BIBentryALTinterwordspacing
M.~Abadi, A.~Agarwal, P.~Barham, E.~Brevdo, Z.~Chen, C.~Citro, G.~S. Corrado,
  A.~Davis, J.~Dean, M.~Devin, S.~Ghemawat, I.~Goodfellow, A.~Harp, G.~Irving,
  M.~Isard, Y.~Jia, R.~Jozefowicz, L.~Kaiser, M.~Kudlur, J.~Levenberg,
  D.~Man\'{e}, R.~Monga, S.~Moore, D.~Murray, C.~Olah, M.~Schuster, J.~Shlens,
  B.~Steiner, I.~Sutskever, K.~Talwar, P.~Tucker, V.~Vanhoucke, V.~Vasudevan,
  F.~Vi\'{e}gas, O.~Vinyals, P.~Warden, M.~Wattenberg, M.~Wicke, Y.~Yu, and
  X.~Zheng, ``{TensorFlow}: Large-scale machine learning on heterogeneous
  systems,'' 2015, software available from tensorflow.org. [Online]. Available:
  \url{https://www.tensorflow.org/}
\BIBentrySTDinterwordspacing

\bibitem{gardner1998artificial}
M.~W. Gardner and S.~Dorling, ``Artificial neural networks (the multilayer
  perceptron)—a review of applications in the atmospheric sciences,''
  \emph{Atmospheric environment}, vol.~32, no. 14-15, pp. 2627--2636, 1998.

\bibitem{kingma2014adam}
D.~P. Kingma and J.~Ba, ``Adam: A method for stochastic optimization,''
  \emph{arXiv preprint arXiv:1412.6980}, 2014.

\bibitem{fscoremeasure}
D.~M.~W. Powers, ``{Evaluation: From precision, recall and f-measure to roc.,
  informedness, markedness \& correlation},'' \emph{Journal of Machine Learning
  Technologies}, vol.~2, no.~1, pp. 37--63, 2011.

\bibitem{Provost98onapplied}
F.~Provost and R.~Kohavi, ``On applied research in machine learning,'' in
  \emph{Machine learning}, 1998, pp. 127--132.

\bibitem{hastie_09}
\BIBentryALTinterwordspacing
T.~Hastie, R.~Tibshirani, and J.~Friedman, \emph{The elements of statistical
  learning: data mining, inference and prediction}, 2nd~ed.\hskip 1em plus
  0.5em minus 0.4em\relax Springer, 2009. [Online]. Available:
  \url{http://www-stat.stanford.edu/~tibs/ElemStatLearn/}
\BIBentrySTDinterwordspacing

\bibitem{kohavi1995study}
R.~Kohavi \emph{et~al.}, ``A study of cross-validation and bootstrap for
  accuracy estimation and model selection,'' in \emph{Ijcai}, vol.~14.\hskip
  1em plus 0.5em minus 0.4em\relax Stanford, CA, 1995, pp. 1137--1145.

\bibitem{NIPS2014_5423}
\BIBentryALTinterwordspacing
I.~Goodfellow, J.~Pouget-Abadie, M.~Mirza, B.~Xu, D.~Warde-Farley, S.~Ozair,
  A.~Courville, and Y.~Bengio, ``Generative adversarial nets,'' in
  \emph{Advances in Neural Information Processing Systems 27}, Z.~Ghahramani,
  M.~Welling, C.~Cortes, N.~D. Lawrence, and K.~Q. Weinberger, Eds.\hskip 1em
  plus 0.5em minus 0.4em\relax Curran Associates, Inc., 2014, pp. 2672--2680.
  [Online]. Available:
  \url{http://papers.nips.cc/paper/5423-generative-adversarial-nets.pdf}
\BIBentrySTDinterwordspacing

\bibitem{wilk1965test}
\BIBentryALTinterwordspacing
M.~B. WILK and S.~S. SHAPIRO, ``{An analysis of variance test for normality
  (complete samples)},'' \emph{Biometrika}, vol.~52, no. 3-4, pp. 591--611, 12
  1965. [Online]. Available: \url{https://doi.org/10.1093/biomet/52.3-4.591}
\BIBentrySTDinterwordspacing

\bibitem{kutner2005applied}
\BIBentryALTinterwordspacing
M.~Kutner, \emph{Applied Linear Statistical Models}, ser. McGrwa-Hill
  international edition.\hskip 1em plus 0.5em minus 0.4em\relax McGraw-Hill
  Irwin, 2005. [Online]. Available:
  \url{https://books.google.com.br/books?id=0xqCAAAACAAJ}
\BIBentrySTDinterwordspacing

\bibitem{schmeling2001age}
A.~Schmeling, A.~Olze, W.~Reisinger, and G.~Geserick, ``Age estimation of
  living people undergoing criminal proceedings,'' \emph{The Lancet}, vol. 358,
  no. 9276, pp. 89--90, 2001.

\bibitem{schmeling2016forensic}
A.~Schmeling, R.~Dettmeyer, E.~Rudolf, V.~Vieth, and G.~Geserick, ``Forensic
  age estimation: methods, certainty, and the law,'' \emph{Deutsches
  {\"A}rzteblatt International}, vol. 113, no.~4, p.~44, 2016.

\bibitem{silva2013evidencia}
R.~F. Silva, S.~D. S.~C. Mendes, A.~F. do~Ros{\'a}rio~J{\'u}nior, P.~E.~M.
  Dias, and L.~B. Martorell, ``Evid{\^e}ncia documental x evid{\^e}ncia
  biol{\'o}gica para estimativa da idade--relato de caso pericial.''
  \emph{Revista Odontol{\'o}gica do Brasil Central}, vol.~22, no.~60, 2013.

\bibitem{machado2017age}
A.~L.~R. Machado, T.~U. Dezem, A.~T. Bruni, and R.~H.~A. da~Silva, ``Age
  estimation by facial analysis based on applications available for
  smartphones,'' \emph{The Journal of forensic odonto-stomatology}, vol.~35,
  no.~2, p.~55, 2017.

\bibitem{deitos2015age}
A.~R. Deitos, C.~Costa, E.~Michel-Crosato, I.~Gali{\'c}, R.~Cameriere, and
  M.~G.~H. Biazevic, ``Age estimation among brazilians: younger or older than
  18?'' \emph{Journal of forensic and legal medicine}, vol.~33, pp. 111--115,
  2015.

\bibitem{santiago2018accuracy}
B.~M. Santiago, L.~Almeida, Y.~W. Cavalcanti, M.~B. Magno, and L.~C. Maia,
  ``Accuracy of the third molar maturity index in assessing the legal age of 18
  years: a systematic review and meta-analysis,'' \emph{International journal
  of legal medicine}, vol. 132, no.~4, pp. 1167--1184, 2018.

\bibitem{franco2013applicability}
A.~Franco, P.~Thevissen, S.~Fieuws, P.~H.~C. Souza, and G.~Willems,
  ``Applicability of willems model for dental age estimations in brazilian
  children,'' \emph{Forensic science international}, vol. 231, no. 1-3, pp.
  401--e1, 2013.

\bibitem{graupner2000sexual}
H.~Graupner, ``Sexual consent: The criminal law in europe and overseas,''
  \emph{Archives of Sexual Behavior}, vol.~29, no.~5, pp. 415--461, 2000.

\bibitem{souza2016science}
V.~S.~d. Souza, ``Science and miscegenation in the early twentieth century:
  Edgard roquette-pinto’s debates and controversies with us physical
  anthropology,'' \emph{Hist{\'o}ria, Ci{\^e}ncias, Sa{\'u}de-Manguinhos},
  vol.~23, no.~3, pp. 597--614, 2016.

\bibitem{edmonds2007triumphant}
A.~Edmonds, ``Triumphant miscegenation: Reflections on beauty and race in
  brazil,'' \emph{Journal of Intercultural Studies}, vol.~28, no.~1, pp.
  83--97, 2007.

\bibitem{kloess2017challenges}
J.~A. Kloess, J.~Woodhams, H.~Whittle, T.~Grant, and C.~E.
  Hamilton-Giachritsis, ``The challenges of identifying and classifying child
  sexual abuse material,'' \emph{Sexual Abuse}, p. 1079063217724768, 2017.

\bibitem{cao2018vggface2}
Q.~Cao, L.~Shen, W.~Xie, O.~M. Parkhi, and A.~Zisserman, ``Vggface2: A dataset
  for recognising faces across pose and age,'' in \emph{2018 13th IEEE
  International Conference on Automatic Face \& Gesture Recognition (FG
  2018)}.\hskip 1em plus 0.5em minus 0.4em\relax IEEE, 2018, pp. 67--74.

\bibitem{xing2017diagnosing}
J.~Xing, K.~Li, W.~Hu, C.~Yuan, and H.~Ling, ``Diagnosing deep learning models
  for high accuracy age estimation from a single image,'' \emph{Pattern
  Recognition}, vol.~66, pp. 106--116, 2017.

\end{thebibliography}

\clearpage
\onecolumn

\appendices
\section{Confusion Matrices}\label{appendix:MAT}

\clearpage
\onecolumn

\section{Photo-anthropometric Indexes}\label{appendix:208_PAI}
\begin{longtable}{lll}
\caption{Description of the 208 photo-anthropometric indexes (PAIs).} \label{tab:long} \\
\hline \multicolumn{1}{l}{\textbf{PAI}} & \multicolumn{1}{l}{\textbf{Landmarks}} & \multicolumn{1}{l}{\textbf{Description}} \\ \hline 
\endfirsthead

\multicolumn{3}{c}%
{{\bfseries \tablename\ \thetable{} -- continued from previous page}} \\
\hline \multicolumn{1}{l}{\textbf{PAI}} & \multicolumn{1}{l}{\textbf{Landmarks}} & \multicolumn{1}{l}{\textbf{Description}} \\ \hline 
\endhead

\hline \multicolumn{3}{r}{{Continued on next page}} \\ \hline
\endfoot

\hline \hline
\endlastfoot
    PAI-0 & al\_r-al\_l $|$ al\_l-al\_r & Nose width \\
    PAI-1 & al\_r-ch\_r $|$ al\_l-ch\_l & Wing of the nose - Labial Commissure (Same side) \\
    PAI-2 & al\_r-ch\_l $|$ al\_l-ch\_r & Wing of the nose - Labial Commissure (Different side) \\
    PAI-3 & al\_r-cph\_r $|$ al\_l-cph\_l & Wing of the nose - Crista philtri (Same side) \\
    PAI-4 & al\_r-cph\_l $|$ al\_l-cph\_r & Wing of the nose - Crista philtri (Different side) \\
    PAI-5 & al\_r-ec\_r $|$ al\_l-ec\_l & Wing of the nose - Ectocanthion (Same side) \\
    PAI-6 & al\_r-ec\_l $|$ al\_l-ec\_r & Wing of the nose - Ectocanthion (Different side) \\
    PAI-7 & al\_r-en\_r $|$ al\_l-en\_l & Wing of the nose  - Endocanthion (Same side)  \\
    PAI-8 & al\_r-en\_l $|$ al\_l-en\_r & Wing of the nose  - Endocanthion (Different side)  \\
    PAI-9 & al\_r-g $|$ al\_l-g & Wing of the nose - Glabella  \\
    PAI-10 & al\_r-gn $|$ al\_l-gn & Wing of the nose - Chin \\
    PAI-11 & al\_r-go\_r $|$ al\_l-go\_l & Wing of the nose - Gonion (Same side) \\
    PAI-12 & al\_r-go\_l $|$ al\_l-go\_r & Wing of the nose - Gonion (Different side) \\
    PAI-13 & al\_r-il\_r $|$ al\_l-il\_l & Wing of the nose - Lateral iris (Same side)  \\
    PAI-14 & al\_r-il\_l $|$ al\_l-il\_r & Wing of the nose - Lateral iris (Different side) \\
    PAI-15 & al\_r-im\_r $|$ al\_l-im\_l & Wing of the nose - Medial iris (Same side) \\
    PAI-16 & al\_r-im\_l $|$ al\_l-im\_r & Wing of the nose - Medial iris (Different side)  \\
    PAI-17 & al\_r-li $|$ al\_l-li & Wing of the nose - Lower lip \\
    PAI-18 & al\_r-ls $|$ al\_l-ls & Wing of the nose - Upper lip  \\
    PAI-19 & al\_r-mid $|$ al\_l-mid & Wing of the nose - Midnasal \\
    PAI-20 & al\_r-n $|$ al\_l-n & Wing of the nose - Nasion \\
    PAI-21 & al\_r-pu\_r $|$ al\_l-pu\_l & Wing of the nose - Pupil (Same side) \\
    PAI-22 & al\_r-pu\_l $|$ al\_l-pu\_r & Wing of the nose - Pupil (Different side) \\
    PAI-23 & al\_r-sn $|$ al\_l-sn & Wing of the nose - Base of the nose  \\
    PAI-24 & al\_r-sto $|$ al\_l-sto & Wing of the nose - Stomion \\
    PAI-25 & al\_r-zy\_r $|$ al\_l-zy\_l & Wing of the nose - Zygion (Same side) \\
    PAI-26 & al\_r-zy\_l $|$ al\_l-zy\_r & Wing of the nose - Zygion (Different side)  \\
    PAI-27 & ch\_r-ch\_l $|$ ch\_l-ch\_r & Mouth width \\
    PAI-28 & ch\_r-cph\_r $|$ ch\_l-cph\_l & Labial Commissure - Crista philtri (Same side) \\
    PAI-29 & ch\_r-cph\_l $|$ ch\_l-cph\_r & Labial Commissure - Crista philtri (Different side) \\
    PAI-30 & ch\_r-ec\_r $|$ ch\_l-ec\_l & Labial Commissure - Ectocanthion (Same side)  \\
    PAI-31 & ch\_r-ec\_l $|$ ch\_l-ec\_r & Labial Commissure - Ectocanthion (Different side)  \\
    PAI-32 & ch\_r-en\_r $|$ ch\_l-en\_l & Labial Commissure - Endocanthion (Same side)  \\
    PAI-33 & ch\_r-en\_l $|$ ch\_l-en\_r & Labial Commissure - Endocanthion (Different side)  \\
    PAI-34 & ch\_r-g $|$ ch\_l-g & Labial Commissure - Glabella  \\
    PAI-35 & ch\_r-gn $|$ ch\_l-gn & Labial Commissure - Chin \\
    PAI-36 & ch\_r-go\_r $|$ ch\_l-go\_l & Labial Commissure - Gonion (Same side) \\
    PAI-37 & ch\_r-go\_l $|$ ch\_l-go\_r & Labial Commissure - Gonion (Different side) \\
    PAI-38 & ch\_r-il\_r $|$ ch\_l-il\_l & Labial Commissure - Lateral iris (Same side) \\
    PAI-39 & ch\_r-il\_l $|$ ch\_l-il\_r & Labial Commissure - Lateral iris (Different side) \\
    PAI-40 & ch\_r-im\_r $|$ ch\_l-im\_l & Labial Commissure - Medial iris (Same side) \\
    PAI-41 & ch\_r-im\_l $|$ ch\_l-im\_r & Labial Commissure - Medial iris (Different side) \\
    PAI-42 & ch\_r-li $|$ ch\_l-li & Labial Commissure - Lower lip \\
    PAI-43 & ch\_r-ls $|$ ch\_l-ls & Labial Commissure - Upper lip \\
    PAI-44 & ch\_r-mid $|$ ch\_l-mid & Labial Commissure - Midnasal \\
    PAI-45 & ch\_r-n $|$ ch\_l-n & Labial Commissure - Nasion \\
    PAI-46 & ch\_r-pu\_r $|$ ch\_l-pu\_l & Labial Commissure - Pupil (Same side) \\
    PAI-47 & ch\_r-pu\_l $|$ ch\_l-pu\_r & Labial Commissure - Pupil (Different side) \\
    PAI-48 & ch\_r-sn $|$ ch\_l-sn & Labial Commissure - Base of the nose \\
    PAI-49 & ch\_r-sto $|$ ch\_l-sto & Labial Commissure - Stomion \\
    PAI-50 & ch\_r-zy\_r $|$ ch\_l-zy\_l & Labial Commissure - Zygion (Same side) \\
    PAI-51 & ch\_r-zy\_l $|$ ch\_l-zy\_r & Labial Commissure - Zygion (Different side) \\
    PAI-52 & cph\_r-cph\_l $|$ cph\_l-cph\_r & Width of the crista philtri \\
    PAI-53 & cph\_r-ec\_r $|$ cph\_l-ec\_l & Crista philtri -  Ectocanthion (Same side) \\
    PAI-54 & cph\_r-ec\_l $|$ cph\_l-ec\_r & Crista philtri -  Ectocanthion (Different side) \\
    PAI-55 & cph\_r-en\_r $|$ cph\_l-en\_l & Crista philtri -  Endocanthion (Same side) \\
    PAI-56 & cph\_r-en\_l $|$ cph\_l-en\_r & Crista philtri -  Endocanthion (Different side) \\
    PAI-57 & cph\_r-g $|$ cph\_l-g & Crista philtri - Glabella  \\
    PAI-58 & cph\_r-gn $|$ cph\_l-gn & Crista philtri - Chin \\
    PAI-59 & cph\_r-go\_r $|$ cph\_l-go\_l & Crista philtri - Gonion (Same side) \\
    PAI-60 & cph\_r-go\_l $|$ cph\_l-go\_r & Crista philtri - Gonion (Different side) \\
    PAI-61 & cph\_r-il\_r $|$ cph\_l-il\_l & Crista philtri - Lateral iris (Same side)  \\
    PAI-62 & cph\_r-il\_l $|$ cph\_l-il\_r & Crista philtri - Lateral iris (Different side)  \\
    PAI-63 & cph\_r-im\_r $|$ cph\_l-im\_l & Crista philtri - Medial iris (Same side)  \\
    PAI-64 & cph\_r-im\_l $|$ cph\_l-im\_r & Crista philtri - Medial iris (Different side)  \\
    PAI-65 & cph\_r-li $|$ cph\_l-li & Crista philtri - Lower lip \\
    PAI-66 & cph\_r-ls $|$ cph\_l-ls & Crista philtri - Upper lip \\
    PAI-67 & cph\_r-mid $|$ cph\_l-mid & Crista philtri - Midnasal  \\
    PAI-68 & cph\_r-n $|$ cph\_l-n & Crista philtri - Nasion \\
    PAI-69 & cph\_r-pu\_r $|$ cph\_l-pu\_l & Crista philtri - Pupil (Same side) \\
    PAI-70 & cph\_r-pu\_l $|$ cph\_l-pu\_r & Crista philtri - Pupil (Different side) \\
    PAI-71 & cph\_r-sn $|$ cph\_l-sn & Crista philtri - Base of the nose \\
    PAI-72 & cph\_r-sto $|$ cph\_l-sto & Crista philtri - Stomion \\
    PAI-73 & cph\_r-zy\_r $|$ cph\_l-zy\_l & Crista philtri - Zygion (Same side) \\
    PAI-74 & cph\_r-zy\_l $|$ cph\_l-zy\_r & Crista philtri - Zygion (Different side) \\
    PAI-75 & ec\_r-ec\_l $|$ ec\_l-ec\_r & Width of the ectocanthion  \\
    PAI-76 & ec\_r-en\_r $|$ ec\_l-en\_l & Eye width  \\
    PAI-77 & ec\_r-en\_l $|$ ec\_l-en\_r & Ectocanthion - Endocanthion (Different side) \\
    PAI-78 & ec\_r-g $|$ ec\_l-g & Ectocanthion - Glabella \\
    PAI-79 & ec\_r-gn $|$ ec\_l-gn & Ectocanthion - Chin \\
    PAI-80 & ec\_r-go\_r $|$ ec\_l-go\_l & Ectocanthion - Gonion (Same side) \\
    PAI-81 & ec\_r-go\_l $|$ ec\_l-go\_r & Ectocanthion - Gonion (Different side) \\
    PAI-82 & ec\_r-il\_r $|$ ec\_l-il\_l & Ectocanthion - Lateral iris (Same side)  \\
    PAI-83 & ec\_r-il\_l $|$ ec\_l-il\_r & Ectocanthion - Lateral iris (Different side)  \\
    PAI-84 & ec\_r-im\_r $|$ ec\_l-im\_l & Ectocanthion - Medial iris (Same side)  \\
    PAI-85 & ec\_r-im\_l $|$ ec\_l-im\_r & Ectocanthion - Medial iris (Different side)  \\
    PAI-86 & ec\_r-li $|$ ec\_l-li & Ectocanthion - Lower lip \\
    PAI-87 & ec\_r-ls $|$ ec\_l-ls & Ectocanthion - Upper lip \\
    PAI-88 & ec\_r-mid $|$ ec\_l-mid & Ectocanthion  - Midnasal \\
    PAI-89 & ec\_r-n $|$ ec\_l-n & Ectocanthion  - Nasion \\
    PAI-90 & ec\_r-pu\_r $|$ ec\_l-pu\_l & Ectocanthion - Pupil (Same side) \\
    PAI-91 & ec\_r-pu\_l $|$ ec\_l-pu\_r & Ectocanthion - Pupil (Different side) \\
    PAI-92 & ec\_r-sn $|$ ec\_l-sn & Ectocanthion - Base of the nose \\
    PAI-93 & ec\_r-sto $|$ ec\_l-sto & Ectocanthion - Stomion \\
    PAI-94 & ec\_r-zy\_r $|$ ec\_l-zy\_l & Ectocanthion - Zygion (Same side) \\
    PAI-95 & ec\_r-zy\_l $|$ ec\_l-zy\_r & Ectocanthion - Zygion (Different side) \\
    PAI-96 & en\_r-en\_l $|$ en\_l-en\_r & Inter-canthion width \\
    PAI-97 & en\_r-g $|$ en\_l-g & Endocanthion - Glabella \\
    PAI-98 & en\_r-gn $|$ en\_l-gn & Endocanthion - Chin \\
    PAI-99 & en\_r-go\_r $|$ en\_l-go\_l & Endocanthion - Gonion (Same side) \\
    PAI-100 & en\_r-go\_l $|$ en\_l-go\_r & Endocanthion - Gonion (Different side) \\
    PAI-101 & en\_r-il\_r $|$ en\_l-il\_l & Endocanthion - Lateral iris (Same side) \\
    PAI-102 & en\_r-il\_l $|$ en\_l-il\_r & Endocanthion - Lateral iris (Different side) \\
    PAI-103 & en\_r-im\_r $|$ en\_l-im\_l & Endocanthion - Medial iris (Same side) \\
    PAI-104 & en\_r-im\_l $|$ en\_l-im\_r & Endocanthion - Medial iris (Different side) \\
    PAI-105 & en\_r-li $|$ en\_l-li & Endocanthion - Lower lip \\
    PAI-106 & en\_r-ls $|$ en\_l-ls & Endocanthion - Upper lip \\
    PAI-107 & en\_r-mid $|$ en\_l-mid & Endocanthion - Midnasal \\
    PAI-108 & en\_r-n $|$ en\_l-n & Endocanthion - Nasion \\
    PAI-109 & en\_r-pu\_r $|$ en\_l-pu\_l & Endocanthion - Pupil (Same side) \\
    PAI-110 & en\_r-pu\_l $|$ en\_l-pu\_r & Endocanthion - Pupil (Different side) \\
    PAI-111 & en\_r-sn $|$ en\_l-sn & Endocanthion - Base of the nose \\
    PAI-112 & en\_r-sto $|$ en\_l-sto & Endocanthion - Stomion \\
    PAI-113 & en\_r-zy\_r $|$ en\_l-zy\_l & Endocanthion - Zygion (Same side) \\
    PAI-114 & en\_r-zy\_l $|$ en\_l-zy\_r & Endocanthion - Zygion (Different side) \\
    PAI-115 & g-gn  & Height of the face \\
    PAI-116 & g-go\_r $|$ g-go\_l & Glabella - Gonion (Same side) \\
    PAI-117 & g-il\_r $|$ g-il\_l & Glabella - Gonion (Different side) \\
    PAI-118 & g-im\_r $|$ g-im\_l & Glabella - Medial iris  \\
    PAI-119 & g-li  & Glabella - Upper lip \\
    PAI-120 & g-ls  & Glabella - Lower lip \\
    PAI-121 & g-mid & Glabella Midnasal  \\
    PAI-122 & g-n   & Glabella - Nasion \\
    PAI-123 & g-pu\_r $|$ g-pu\_l & Glabella - Pupil \\
    PAI-124 & g-sn  & Glabella - Base of the nose \\
    PAI-125 & g-sto & Glabella - Stomion \\
    PAI-126 & g-zy\_r $|$ g-zy\_l & Glabella - Zygion \\
    PAI-127 & gn-go\_r $|$ gn-go\_l & Chin - Gonion \\
    PAI-128 & gn-il\_r $|$ gn-il\_l & Chin - Lateral iris \\
    PAI-129 & gn-im\_r $|$ gn-im\_l & Chin - Medial iris  \\
    PAI-130 & gn-li & Chin - Lower lip \\
    PAI-131 & gn-ls & Chin - Upper lip \\
    PAI-132 & gn-mid & Chin - Midnasal \\
    PAI-133 & gn-n  & Chin - Nasion \\
    PAI-134 & gn-pu\_r $|$ gn-pu\_l & Chin - Pupil \\
    PAI-135 & gn-sn & Chin - Base of the nose \\
    PAI-136 & gn-sto & Chin - Stomion \\
    PAI-137 & gn-zy\_r $|$ gn-zy\_l & Chin - Zygion \\
    PAI-138 & go\_r-go\_l $|$ go\_l-go\_r & Inter-gonion width \\
    PAI-139 & go\_r-il\_r $|$ go\_l-il\_l & Gonion - Lateral iris (Same side)  \\
    PAI-140 & go\_r-il\_l $|$ go\_l-il\_r & Gonion - Lateral iris (Different side) \\
    PAI-141 & go\_r-im\_r $|$ go\_l-im\_l & Gonion - Medial iris (Same side)  \\
    PAI-142 & go\_r-im\_l $|$ go\_l-im\_r & Gonion - Medial iris (Different side) \\
    PAI-143 & go\_r-li $|$ go\_l-li & Gonion - Lower lip \\
    PAI-144 & go\_r-ls $|$ go\_l-ls & Gonion - Upper lip \\
    PAI-145 & go\_r-mid $|$ go\_l-mid & Gonion - Midnasal \\
    PAI-146 & go\_r-n $|$ go\_l-n & Gonion - Nasion \\
    PAI-147 & go\_r-pu\_r $|$ go\_l-pu\_l & Gonion - Pupil (Same side)  \\
    PAI-148 & go\_r-pu\_l $|$ go\_l-pu\_r & Gonion - Pupil (Different side)  \\
    PAI-149 & go\_r-sn $|$ go\_l-sn & Gonion - Base of the nose \\
    PAI-150 & go\_r-sto $|$ go\_l-sto & Gonion - Stomion \\
    PAI-151 & go\_r-zy\_r $|$ go\_l-zy\_l & Gonion - Zygion (Same side) \\
    PAI-152 & go\_r-zy\_l $|$ go\_l-zy\_r & Gonion - Zygion (Different side) \\
    PAI-153 & il\_r-il\_l $|$ il\_l-il\_r & Maximum iris width \\
    PAI-154 & il\_r-im\_r $|$ il\_l-im\_l & Diameter of the iris \\
    PAI-155 & il\_r-im\_l $|$ il\_l-im\_r & Lateral iris - Medial iris (Different side) \\
    PAI-156 & il\_r-li $|$ il\_l-li & Lateral iris - Upper lip \\
    PAI-157 & il\_r-ls $|$ il\_l-ls & Lateral iris - Upper lip \\
    PAI-158 & il\_r-mid $|$ il\_l-mid & Lateral iris - Midnasal \\
    PAI-159 & il\_r-n $|$ il\_l-n & Lateral iris - Nasion \\
    PAI-160 & il\_r-pu\_r $|$ il\_l-pu\_l & Lateral iris - Pupil (Same side)  \\
    PAI-161 & il\_r-pu\_l $|$ il\_l-pu\_r & Lateral iris - Pupil (Different side)  \\
    PAI-162 & il\_r-sn $|$ il\_l-sn & Lateral iris - Base of the nose \\
    PAI-163 & il\_r-sto $|$ il\_l-sto & Lateral iris - Stomion \\
    PAI-164 & il\_r-zy\_r $|$ il\_l-zy\_l & Lateral iris - Zygion (Same side) \\
    PAI-165 & il\_r-zy\_l $|$ il\_l-zy\_r & Lateral iris - Zygion (Different side) \\
    PAI-166 & im\_r-im\_l $|$ im\_l-im\_r & Minimum iris width  \\
    PAI-167 & im\_r-li $|$ im\_l-li & Medial iris - Lower lip \\
    PAI-168 & im\_r-ls $|$ im\_l-ls & Medial iris - Upper lip \\
    PAI-169 & im\_r-mid $|$ im\_l-mid & Medial iris - Midnasal \\
    PAI-170 & im\_r-n $|$ im\_l-n & Medial iris - Nasion \\
    PAI-171 & im\_r-pu\_r $|$ im\_l-pu\_l & Medial iris - Pupil (Same side) \\
    PAI-172 & im\_r-pu\_l $|$ im\_l-pu\_r & Medial iris - Pupil (Different side) \\
    PAI-173 & im\_r-sn $|$ im\_l-sn & Medial iris - Base of the nose \\
    PAI-174 & im\_r-sto $|$ im\_l-sto & Medial iris - Stomion \\
    PAI-175 & im\_r-zy\_r $|$ im\_l-zy\_l & Medial iris - Zygion (Same side) \\
    PAI-176 & im\_r-zy\_l $|$ im\_l-zy\_r & Medial iris - Zygion (Different side) \\
    PAI-177 & li-ls & Lip height \\
    PAI-178 & li-mid & Lower lip - Midnasal \\
    PAI-179 & li-n  & Lower lip - Nasion \\
    PAI-180 & li-pu\_r $|$ li-pu\_l & Lower lip - Pupil \\
    PAI-181 & li-sn & Lower lip - Base of the nose  \\
    PAI-182 & li-sto & Lower lip - Stomion \\
    PAI-183 & li-zy\_r $|$ li-zy\_l & Lower lip - Zygion \\
    PAI-184 & ls-mid & Upper lip - Midnasal \\
    PAI-185 & ls-n  & Upper lip - Nasion \\
    PAI-186 & ls-pu\_r $|$ ls-pu\_l & Upper lip - Pupil \\
    PAI-187 & ls-sn & Upper lip - Base of the nose  \\
    PAI-188 & ls-sto & Upper lip - Stomion \\
    PAI-189 & ls-zy\_r $|$ ls-zy\_l & Upper lip - Zygion  \\
    PAI-190 & mid-n & Midnasal - Nasion \\
    PAI-191 & mid-pu\_r $|$ mid-pu\_l & Midnasal - Pupil \\
    PAI-192 & mid-sn & Midnasal - Base of the nose \\
    PAI-193 & mid-sto & Midnasal - Stomion  \\
    PAI-194 & mid-zy\_r $|$ mid-zy\_l & Midnasal - Zygion \\
    PAI-195 & n-pu\_r $|$ n-pu\_l & Nasion - Pupil \\
    PAI-196 & n-sn  & Nose height \\
    PAI-197 & n-sto & Nasion - Stomion \\
    PAI-198 & n-zy\_r $|$ n-zy\_l & Nasion - Zygion \\
    PAI-199 & pu\_r-pu\_l $|$ pu\_l-pu\_r & Inter-pupil width \\
    PAI-200 & pu\_r-sn $|$ pu\_l-sn & Pupil - Base of the nose \\
    PAI-201 & pu\_r-sto $|$ pu\_l-sto & Pupil - Stomion \\
    PAI-202 & pu\_r-zy\_r $|$ pu\_l-zy\_l & Pupil - Zygion (Same side) \\
    PAI-203 & pu\_r-zy\_l $|$ pu\_l-zy\_r & Pupil - Zygion (Different side) \\
    PAI-204 & sn-sto & Base of the nose - Stomion \\
    PAI-205 & sn-zy\_r $|$ sn-zy\_l & Base of the nose - Zygion \\
    PAI-206 & sto-zy\_r $|$ sto-zy\_l & Stomion - Zygion \\
    PAI-207 & zy\_r-zy\_l $|$ zy\_l-zy\_r & Face width \\
\end{longtable}
\clearpage
\twocolumn




%








\end{document}